%% file: manuscript.tex
\documentclass[11pt]{article}

\usepackage[utf8]{inputenc}
\usepackage[T1]{fontenc}
\usepackage{amsmath}
\usepackage{booktabs}
\usepackage{graphicx}
\usepackage{float}
\usepackage{url}
\usepackage[margin=1in]{geometry}
\usepackage[numbers,sort&compress]{natbib}
\usepackage{setspace}
\usepackage{caption}
\usepackage{authblk}
\newcommand{\dv}{\ensuremath{\Delta V_{24}}}

\title{Causal-prior-constrained deep learning for rapid-intensification
warning of western North Pacific tropical cyclones}
\author[a]{Xinrui Li}
\author[a]{Yijun Jin}
\author[b]{Yifan Dong}
\author[a]{Shoujuan Shu \thanks{Corresponding author: sjshu@zju.edu.cn}}
\affil[a]{Department of Atmospheric Sciences, School of Earth Sciences, Zhejiang University, Hangzhou, 310058, China}
\affil[b]{School of Mathematical Sciences, Zhejiang University, Hangzhou, 310058, China}
\date{October 28, 2026}

\begin{document}
\maketitle

% ============================================================ ABSTRACT
\begin{abstract}
Rapid intensification (RI) is one of the most dangerous and unpredictable
phases of tropical cyclones, and its environmental and structural drivers are
difficult to identify and separate in observations. Here we introduce CausalVisionRI, a
causal-prior-constrained deep-learning system for 24-h RI warning in the
western North Pacific. It integrates infrared remote sensing imagery,
atmospheric and oceanic fields, and an observationally discovered causal
graph into a unified model. The graph and eight sign-verified effects on
24-h intensity change provide an auditable causal prior for training.
Discovered from 522 tropical cyclones in 2000--2017, it identifies causal pathways
linking atmospheric and oceanic conditions, the cyclone's dynamical and
structural state, and its 24-h intensity change. On a forward holdout
covering 2020--2024 and 3{,}004 records, the five-member ensemble achieves
a Brier skill score (BSS) of 0.259 relative to a fixed development
climatology, significantly exceeding four reference systems; the three
LDA baselines reach at most 0.105. It also attains higher detection rates
than a published knowledge-block scheme and a reconstructed contrastive-learning
scheme at matched false-detection rates under our USA 30-kt standard and the
published RITCF standard. Ablations show that the graph pathway adds 0.032
BSS across the holdout and 0.063 over 2020--2021, indicating that the gain
concentrates in the time period adjacent to the estimation window. 
A deep-learning model constrained by observationally discovered causal structure 
thus offers a new route to physically informed RI warning.
\end{abstract}

% ============================================================ INTRO
\section{Introduction}

Rapid intensification (RI) of tropical cyclones (TCs), defined by an increase
of at least 30~kt in maximum sustained wind within 24~h \citep{kaplan2003ri,kaplan2010rii}, is one of the most
hazardous stages of TC development and can cause severe disasters and
socio-economic impacts when TCs make landfall. Its
predictability is low because it is triggered by complex interacting
oceanic and atmospheric processes that are difficult to identify and
separate in observational data. Numerical experiments reveal these processes
only at considerable computational cost and with few samples of rare events.
As a result, RI is associated with large errors in operational intensity
forecasts \citep{kaplan2010rii,trabing2020errors}. It can also reduce the time
available for officials to issue evacuation orders \citep{greer2024dual}.
Two approaches to RI prediction therefore coexist in practice: statistical and
statistical--dynamical schemes built on hand-selected
predictors \citep{demaria2005ships,kaplan2010rii,horinouchi2025wrps1}, and
deep-learning systems that learn from satellite imagery and environmental
fields directly \citep{wang2025ritcf}.

Three challenges motivate this work. Deep-learning RI evaluation has relied
mainly on detection metrics such as probability of detection and false-alarm
ratio. Proper scoring rules and statistical--dynamical baselines are seldom
reported together, limiting comparisons of forecast-probability quality across
studies. Deep-learning RI prediction also needs explicit, checkable accounts
of the environmental couplings behind its predictions. Causal inference on
Earth-system time series provides a practical toolbox for this purpose
\citep{runge2019pcmci,runge2020pcmcip}, and causal discovery has been used to
select predictors for statistical TC intensity models
\citep{ganesh2025multidata}. Here the complete discovered graph and its signed
effects form a deep-learning prior for RI. Finally, published systems are
verified against different label sources, in different basins and over
different time windows, which leaves the reported numbers incomparable as they
stand \citep{wang2025ritcf,horinouchi2025wrps1,eusebi2025salinity}.

In this paper we identify causal pathways of 24-h intensity change underlying
RI, and combine a causal prior with infrared remote sensing imagery and
atmospheric and oceanic fields in a unified deep-learning model
for 24-h RI warning in the western North Pacific (WNP). The model,
CausalVisionRI, blends these three sources and constrains training with the
causal graph and eight sign-verified effects on 24-h intensity change as an
auditable prior. The graph, discovered from 522 TCs over 2000--2017,
links atmospheric and oceanic conditions to the cyclone's dynamical and
structural state, and through that state to 24-h intensity change. On a
forward holdout of 3{,}004 rows from 2020--2024,
evaluated once after the protocol lock, the equal-weight five-member ensemble
(CausalVisionRI-Ensemble, CVRI-Ens) reaches a Brier skill score (BSS) of 0.259
relative to a fixed development climatology. It significantly exceeds four
reference systems under the identical protocol, and at matched false-detection
rates it outperforms both published schemes as reconstructed here under our USA 30-kt standard and
the published RITCF CMA 25.27-kt standard. The protocol-locked ablation shows
that the causal prior improves overall model performance primarily through
the graph pathway, with the largest ensemble gains in the two test years
nearest the estimation window. To our knowledge, CausalVisionRI is the first
RI warning system to encode an observationally discovered causal graph and
its sign-verified effects as a unified training prior. The accompanying
evaluation is also the first in the WNP to report proper skill scores
alongside statistical--dynamical baselines under a common, strictly forward
holdout protocol.

% ============================================================ RESULTS
\section{Results}

\subsection{Causal discovery on 522 observed TCs identifies the causal chain of 24-h intensity change}

\begin{figure}[!htbp]
\centering
\includegraphics[width=\textwidth]{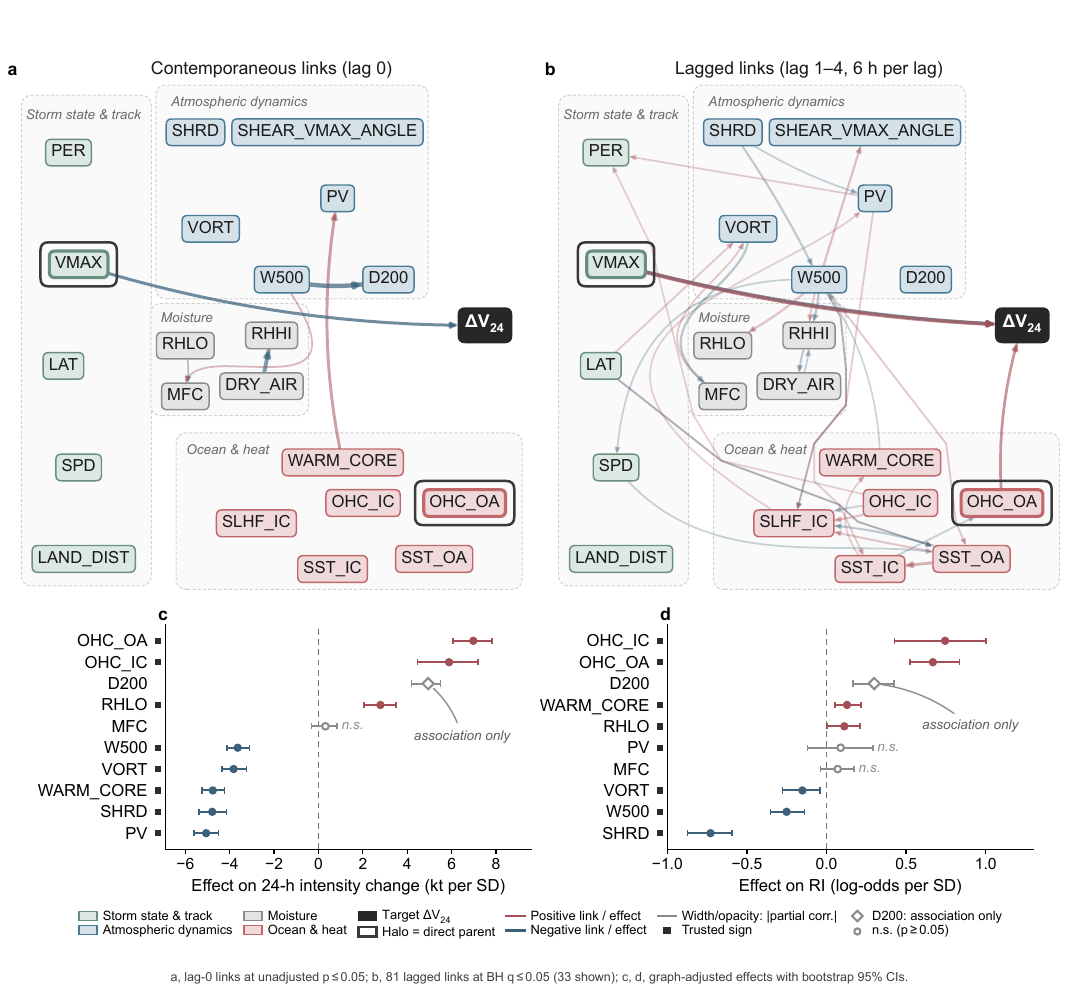}
\caption{\textbf{The causal graph discovered from 522 observed
TCs in 2000--2017.} \textbf{a},\textbf{b}, Prior-constrained PCMCI+
\citep{runge2020pcmcip} graph (fit 2000--2017, $n=12{,}647$): \textbf{a},
six contemporaneous (lag 0) links at unadjusted $p\le0.05$; \textbf{b}, 81
lagged links at Benjamini--Hochberg false discovery rate (FDR) $q\le0.05$
(6~h per lag). For
legibility panel \textbf{b} retains the 32 strongest lagged partial
correlations together with every direct parent of the
24-h intensity change \dv\ (33 of 81 significant lagged links shown); node
fill denotes the variable family, edge colour the sign of the partial
correlation and edge width/opacity its magnitude. Node outlines are
doubled (halo) for the two direct parents of \dv: current intensity (VMAX)
and outer-area ocean heat content (OHC\_OA). \textbf{c},\textbf{d},
Graph-adjusted effect estimates with bootstrap 95\% confidence
intervals (CIs; $B=200$) on 24-h intensity change (\textbf{c}) and on
rapid intensification (RI) as conditional log-odds per SD (\textbf{d}), ordered by point
estimate; solid black squares mark the eight treatments whose
bootstrap-stable \dv\ signs enter the causal training loss, the open diamond
marks upper-level divergence D200 (association only on both scales; no
directed path to \dv\ in the identification DAG, with an inserted edge for
its RI estimate) and open
grey circles are not significant (bootstrap $p\ge0.05$).}
\label{fig:discovery}
\end{figure}

We use masked, prior-constrained PCMCI+ \citep{runge2020pcmcip} to discover
causal links from 12{,}647 six-hourly observations of 522 TCs in
2000--2017. The process generates a graph with 87 directed links
at a significance level of 0.05, consisting of 81 lagged links at
Benjamini--Hochberg false discovery rate (FDR) $q\le0.05$ and six
contemporaneous links at unadjusted $p\le0.05$
(Fig.~\ref{fig:discovery}a,b). In the discovered graph, current intensity
(VMAX; lag-1 partial correlation $+0.150$, plus direct links at lag 0 and
lag 3) and outer-area ocean heat content (OHC\_OA; lag-1 $+0.062$,
$q=1.8\times10^{-7}$; variable definitions in Table~\ref{tab:variables})
are the only two incoming neighbors of the 24-h intensity-change target.
The lag-0 VMAX link is definitional, because
\dv\ contains $-$VMAX($t$) by construction; only the lagged links carry
predictive information. Every other environmental variable with a directed
path to the target reaches it indirectly, through storm-state and ocean nodes.
Graph-adjusted effect estimates (Fig.~\ref{fig:discovery}c; full set in Supplementary Table S1) rank both
OHC terms highest among the positive effects. OHC\_OA has the largest
identified effect on \dv: $+6.98$~kt per 24~h per standard deviation (SD).
Its bootstrap 95\% confidence interval (CI) is [6.06, 7.82], conditional on
the locked identification directed acyclic graph (DAG). Potential vorticity
(PV; $-5.06$), deep-layer shear (SHRD; $-4.79$), warm-core anomaly
(WARM\_CORE; $-4.76$), low-level vorticity (VORT; $-3.83$) and mid-level
pressure velocity (W500; $-3.64$; negative W500 values denote ascent) have
negative estimated effects on \dv. These effects describe the defined \dv\ outcome,
including its negative current-intensity term. On the RI outcome (conditional
log-odds scale), six of the eight treatments that pass the checks below
keep their \dv\ sign; the warm-core anomaly is positive and significant
for RI ($+0.13$), and PV is positive but not significant
(Fig.~\ref{fig:discovery}d).

We use identification and refutation checks to select the effect signs that
enter the model as constraints (discovery assumptions and their consequences
are summarized in Supplementary Table S10). Moisture flux convergence (MFC) has no detectable
effect (bootstrap $p=0.31$). Upper-level divergence (D200) shows a
significant \dv-scale association ($+4.94$), has no directed path to \dv\
in the identification DAG, and reaches RI only through an inserted structural
edge. Both estimates are therefore reported as associations, and D200 is
excluded from the causal ranking.
Eight treatments pass the bootstrap, refutation and directed-path checks and
provide the trusted \dv\ signs used in training: SHRD$-$, RHLO$+$, VORT$-$,
PV$-$, WARM\_CORE$-$, W500$-$, OHC\_IC$+$, OHC\_OA$+$
(Fig.~\ref{fig:discovery}c,d, black squares).
The trusted signs remain stable across admissible constructions of the
identification DAG. For OHC\_OA, the estimated
effect ranges from $+3.39$ to $+8.92$ across these constructions
(Supplementary Note S1).
The trusted effects are essentially unchanged under the strict lag-$\geq$1
configuration: OHC\_OA is $+7.07$, versus $+6.98$ at baseline. Its lag-1 link
into the target persists; PER enters at lag 1, and VMAX and D200 enter at
lag 2. The Madden--Julian oscillation (MJO) and the El Ni\~no--Southern
Oscillation (ENSO) modulate large-scale convection and circulation, as
well as TC activity and intensity in the western North Pacific
\citep{klotzbach2014mjo,camargo2005enso}. Because these climate modes can
affect both the environmental treatments and intensity change, they are
candidate confounders for the observational estimates. Adding the
MJO index (OMI1/OMI2) \citep{kiladis2014omi} and Ni\~no-3.4 indices to
the adjustment sets leaves every sign unchanged (OHC\_OA $+6.98 \to +7.03$;
all other \dv-scale shifts within 4.4\%). The trusted sign structure is
therefore insensitive to these large-scale climate-state adjustments.
Adding latitude to the OHC\_OA adjustment set leaves the effect large
($+6.48$). Full sensitivity
results are reported in Supplementary Note S2.

\subsection{CausalVisionRI embeds the discovered causal graph into
multi-source RI prediction}

\begin{figure}[!htbp]
\centering
\includegraphics[width=\textwidth]{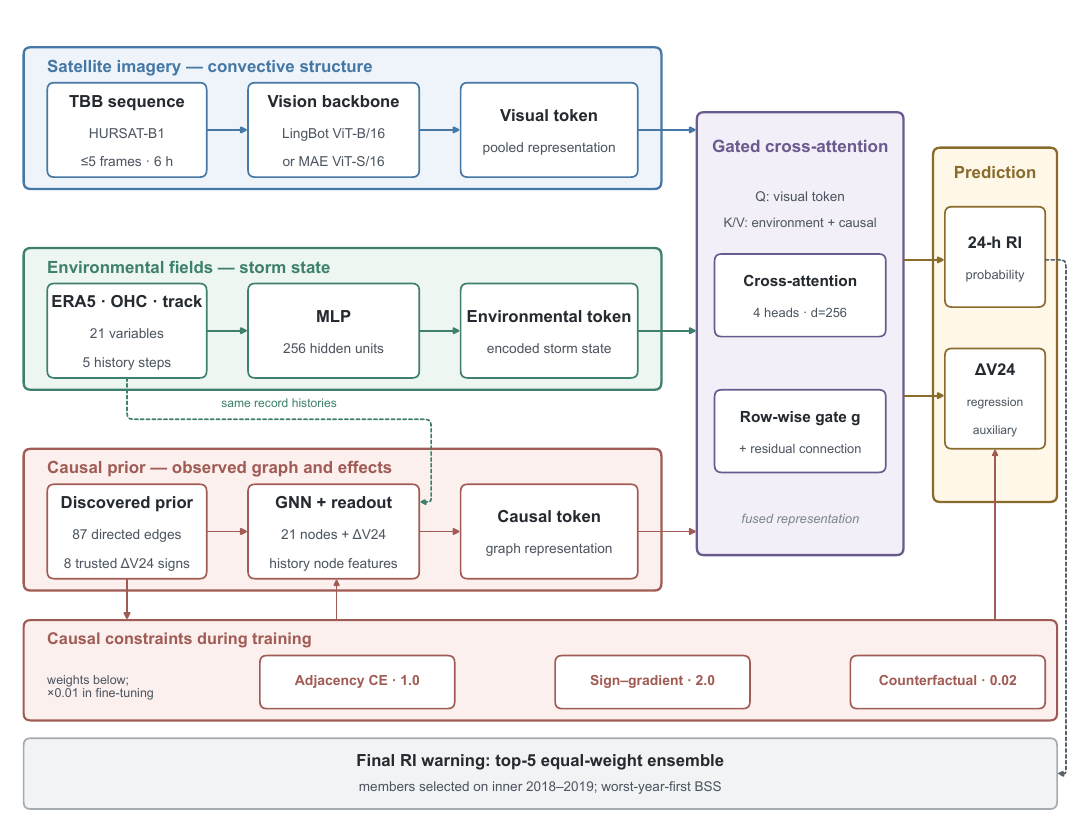}
\caption{\textbf{CausalVisionRI fuses satellite, environmental and
causal-graph streams under sign-verified constraints.} One branch reads
storm-centred infrared imagery (TBB, black-body brightness temperature)
with a ViT-B/16 vision-transformer backbone initialised from the
LingBot-Vision base student \citep{fu2026lingbotvision}. A second condenses the 21-variable
environmental state over five history steps through a
multilayer-perceptron (MLP) encoder. These per-record histories also supply
the GNN node features, while the discovered graph supplies its fixed
topology and edge attributes. The pooled visual token attends to the
environmental and causal tokens, with a per-record sigmoid gate $g$ scaling
the attention output. Adjacency cross-entropy matches incoming-edge
attention to the coefficient-normalized prior weights. Sign--gradient and
counterfactual losses use the eight estimated effect signs to constrain the
\dv\ environmental response. The selected system is the equal-weight ensemble of
five members selected on inner-period (2018--2019) skill; configuration
details are given in the Methods.}
\label{fig:architecture}
\end{figure}

CausalVisionRI (Fig.~\ref{fig:architecture}) is a causal-informed
multimodal model that blends a causal prior, infrared remote sensing imagery,
and atmospheric and oceanic fields at each synoptic time. The satellite branch reads the
convective structure from black-body brightness temperature (TBB)
sequences, with a ViT-B/16 vision transformer initialised from the
LingBot-Vision base student \citep{fu2026lingbotvision}. The
environmental branch condenses 21 variables over five 6-h steps with a
multilayer perceptron (MLP). The same per-record histories supply node
features to the graph neural network (GNN), while the discovered graph
supplies its fixed topology and edge attributes (87 edges carrying lag,
sign and coefficient magnitude). A gated
cross-attention block fuses the three streams: the pooled visual token
queries the environmental and causal tokens, and a per-record sigmoid gate
scales the attention output. Three loss terms link the training
objective to the prior: an
adjacency term that matches incoming-edge attention to the prior
coefficient weights, a sign term that penalizes violations of the eight
trusted \dv\ signs, and a counterfactual term that penalizes intensity
predictions that increase when trusted environmental inputs are shifted
against their effect signs.

We trained 22 development arms on fit-period data and used only inner-period
data for model selection. The locked rule prioritizes the worst-year inner BSS
to favour robust performance across inner years. A family champion receives a
protected slot only when it ranks within the overall top eight; the remaining
slots are filled by overall rank. Both MAE champions fall outside that gate,
leaving five LingBot members. Their equal-weight ensemble raises the pooled
inner BSS from 0.316 for the highest-scoring single arm to 0.332 (all 22 arms
in Supplementary Table S2 and Supplementary Fig.~S1). All test-period results
were computed once from frozen predictions after the final protocol lock.

\subsection{CVRI-Ens improves probabilistic RI prediction in the
2020--2024 holdout}

\begin{figure}[!htbp]
\centering
\includegraphics[width=\textwidth]{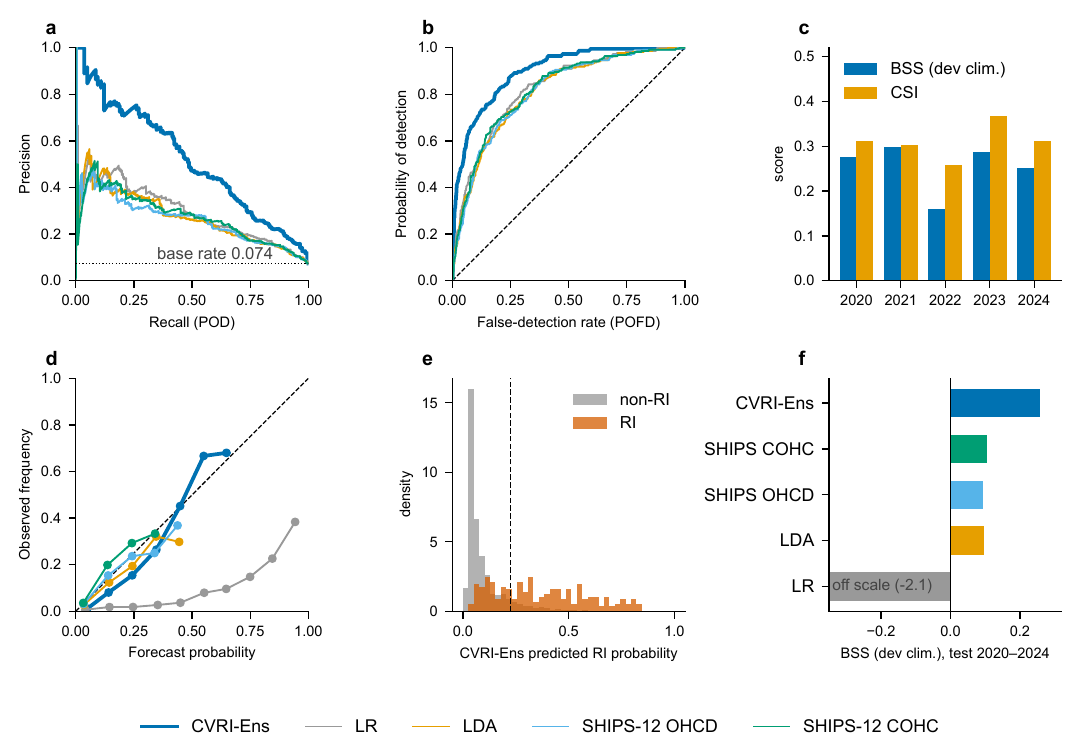}
\caption{\textbf{CVRI-Ens attains a Brier skill score (BSS) of 0.259 in
the 2020--2024 holdout, exceeding all four reference systems.}
\textbf{a}, precision--recall curves (dashed line: rapid
intensification (RI) base rate 7.39\%); \textbf{b},
receiver-operating-characteristic curves; \textbf{c}, per-year BSS
(development climatology) and critical success index (CSI); \textbf{d},
reliability diagram; \textbf{e}, predicted-probability distributions for
RI and non-RI records (dashed line: inner-derived threshold 0.225);
\textbf{f}, BSS of CVRI-Ens (CausalVisionRI-Ensemble) against the four
reference systems (logistic regression, LR, is off scale). References
share the same rows and threshold rule throughout.}
\label{fig:main}
\end{figure}

We evaluate CVRI-Ens on 3{,}004 records from 131 TCs in the
2020--2024 holdout, with an RI base rate of 7.39\%.
CVRI-Ens attains a Brier skill score against the fixed 2000--2019
development climatology ($\mathrm{BSS_{dev}}$) of 0.259
(Table~\ref{tab:main}, Fig.~\ref{fig:main}), with an area under the
receiver-operating-characteristic curve (ROC-AUC) of 0.900 and an area
under the precision--recall curve (PR-AUC) of 0.518 at the same base
rate. At the inner-derived threshold (0.225), the ensemble has a
probability of detection (POD) of 0.680 at a probability of false
detection (POFD) of 0.095 (false-alarm ratio, FARatio, 0.635; critical
success index, CSI, 0.311). Skill is positive in every test year (per-year metrics in Supplementary Table S3), from
0.159 in 2022 to 0.298 in 2021. The five members individually reach
0.151--0.214, so the equal-weight ensemble exceeds every member
(Supplementary Fig.~S2). Sensitivity ensembles score 0.259 for K4 and
Top-5, 0.279 for K6 and 0.244 for all-22 (Supplementary Table S4);
Top-5 remains the protocol-locked primary ensemble.

The three LDA references reach BSS values of 0.095--0.105 under the
identical protocol (Table~\ref{tab:main}): the 12-predictor Statistical
Hurricane Intensity Prediction Scheme (SHIPS-12) linear discriminant
analysis (LDA) reaches 0.105 with climatological OHC (COHC) and 0.095 with
time-dependent daily OHC (OHCD), and the 21-predictor panel LDA reaches
0.098. Class-balanced logistic regression (LR) provides an uncalibrated
reference. Storm-cluster paired bootstrap differences relative to CVRI-Ens
are positive and Holm-significant for every reference ($\Delta$BSS $+0.154$
to $+2.333$, $p\le0.003$; 90\% CIs in Table~\ref{tab:main}). The ensemble's
threshold-free discrimination (ROC-AUC 0.900) exceeds the best reference
(0.826). The ensemble's precision--recall curve lies at or above the reference
curves for recall $>0.1$ in the full holdout (Fig.~\ref{fig:main}a) and for
recall $>0.14$ in the shared comparison sample (Supplementary Fig.~S3).
Its receiver-operating-characteristic curve lies at or above the reference curves
throughout in both comparisons (Fig.~\ref{fig:main}b and Supplementary
Fig.~S3). The lowest probability bin $[0,0.1)$ contains 2{,}019 of the
3{,}004 records (67.2\%; reliability data in \texttt{main\_2024.json}), with
mean prediction 0.049 against an observed frequency of 0.011
(Fig.~\ref{fig:main}d). The threshold (0.225) lies above this bin
(Fig.~\ref{fig:main}e), so the thresholded metrics are unaffected.

\begin{table}[!htbp]
\centering
\caption{\textbf{CVRI-Ens attains the highest BSS, ROC-AUC, PR-AUC and CSI,
and the lowest FARatio, in the 2020--2024 holdout ($n=3{,}004$ rows,
131 systems; RI base rate 7.39\%).}
All models share the same rows and metrics; operating metrics use each
model's own inner-period F1 threshold (F1, harmonic mean of precision and recall).
$\mathrm{BSS_{dev}}$, Brier skill score against the fixed
2000--2019 development climatology (0.0791). $\Delta$BSS is the
storm-cluster paired bootstrap difference relative to CVRI-Ens, Holm
adjusted; 90\% confidence intervals (CIs) in brackets. POD, probability
of detection; POFD, probability of false detection; FARatio, false-alarm
ratio ($1-$precision); CSI, critical success index; ROC-AUC and PR-AUC,
areas under the receiver-operating-characteristic and precision--recall
curves; LR, logistic regression; LDA, linear discriminant analysis;
SHIPS-12, 12-predictor Statistical Hurricane Intensity Prediction Scheme;
COHC/OHCD, climatological ocean heat content / time-dependent daily OHC. Best value
in each metric column in bold.}
\label{tab:main}
\scriptsize
\setlength{\tabcolsep}{4pt}
\begin{tabular}{lcccccccc}
\toprule
Model & BSS$_{\mathrm{dev}}$ & ROC-AUC & PR-AUC & POD & POFD & FARatio & CSI & $\Delta$BSS [90\% CI] \\
\midrule
\textbf{CVRI-Ens} & \textbf{0.259} & \textbf{0.900} & \textbf{0.518} &
0.680 & 0.095 & \textbf{0.635} & \textbf{0.311} & -- \\
SHIPS-12 LDA (COHC) & 0.105 & 0.820 & 0.276 & 0.347 & \textbf{0.060} & 0.686 &
0.197 & $+0.154$ [$+0.079$, $+0.214$] \\
SHIPS-12 LDA (OHCD) & 0.095 & 0.813 & 0.266 & 0.473 & 0.095 & 0.715 &
0.216 & $+0.164$ [$+0.095$, $+0.222$] \\
LDA (21 panel predictors) & 0.098 & 0.813 & 0.278 & \textbf{0.730} & 0.250 &
0.811 & 0.176 & $+0.161$ [$+0.097$, $+0.215$] \\
LR (uncalibrated) & $-2.074$ & 0.826 & 0.298 & 0.649 & 0.173 & 0.769 & 0.205 &
$+2.333$ [$+1.935$, $+2.866$] \\
\bottomrule
\end{tabular}
\end{table}

\subsection{The graph pathway adds holdout skill, concentrated near its
estimation window}

The protocol-locked member-level comparison with C2, which removes the GNN
branch and disables the causal losses, is positive in three of five
members and Holm-significant in LoRA-sf ($\Delta$BSS $+0.080$,
$p_{\mathrm{Holm}}=0.004$); the other four have intervals that include zero
($\Delta$BSS $-0.026$ to $+0.032$, $p_{\mathrm{Holm}}\ge0.43$;
Supplementary Table S5). Over the full test period, all eight LoRA-sf
ablation contrasts are positive, six of them significant.

\begin{figure}[!htbp]
\centering
\includegraphics[width=\textwidth]{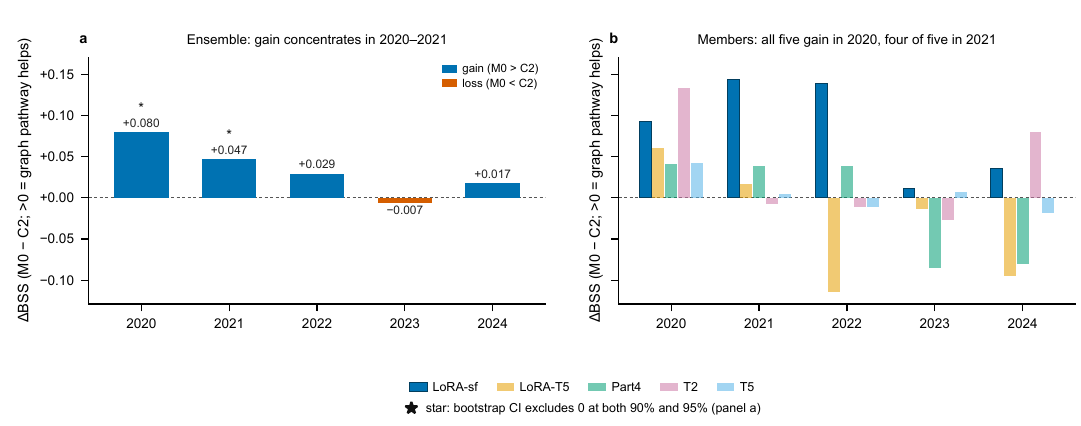}
\caption{\textbf{Disabling the graph pathway costs significant holdout
skill in 2020 and 2021.} \textbf{a}, Per-year change in
Brier skill score of the ensemble when the graph pathway is disabled
(C2); positive values favour the graph pathway, and stars mark years where the
gain is significant under the locked bootstrap rule. \textbf{b}, The same
contrast for the five members individually; all five members gain in
2020 and four of five in 2021. In panel \textbf{a}, blue and orange
distinguish positive and negative changes; member colours in panel \textbf{b}
identify the ensemble members.}
\label{fig:ablation}
\end{figure}

Removing the graph pathway lowers the ensemble's pooled BSS 
by 0.032 across all 3,004 holdout rows in 2020--2024. 
The year-by-year changes show that both years nearest the estimation window show
significant gains: $+0.080$ in 2020 (90\% CI [$+0.043$, $+0.140$],
$p<0.001$) and $+0.047$ in 2021 ([$+0.023$, $+0.076$], $p=0.003$), with
$+0.063$ ([$+0.042$, $+0.090$], $p<0.001$) for the pooled two-year period
(Fig.~\ref{fig:ablation}a for the annual contrasts;
\texttt{year\_gradient\_ablation.json} for the pooled values).
All five members improve in 2020 and four of five improve in 2021
(Fig.~\ref{fig:ablation}b). From 2022 onward the ensemble contrast is smaller
and its interval includes zero.

Two control families separate the graph pathway from the auxiliary losses
and from generic capacity. Removing all three causal losses while retaining
the branch (C1) leaves the ensemble's 2020--2021 skill statistically
unchanged ($\Delta$BSS $+0.007$, 90\% CI [$-0.013$, $+0.026$]). Over the
full test period, replacing the prior graph with shuffles that preserve each node's degree in the
environmental subgraph (C3), with input dimensionality and loss weights
unchanged, lowers LoRA-sf skill under all three seeds (by up to 0.062, two
significantly). For LoRA-sf, this gain is sensitive to the discovered
topology. Across the full 40-contrast panel (Supplementary Table S5), ten
contrasts survive Holm correction and seven favour the prior. The three
significant negative contrasts occur in Part4 (C1 and C4sign) and T2 (C4sign),
showing that the full constraint set has member-specific effects while
improving the overall skill of the final model.

The members also use the prior differently. Part4 routes 69\% of its
context attention to the causal token, whereas LoRA-sf routes 23\% yet shows
the largest graph-removal effect (Supplementary Table S5). Attention share and ablation benefit
capture different aspects of how the members use the prior (Supplementary
Fig.~S6 and Table~S6).
The 2020--2021 member panel is shown in Supplementary Fig.~S5; per-contrast
and per-year ablation values are given in the released analysis files.

\subsection{CVRI-Ens improves detection at matched false-detection rates
under common evaluation standards}

CVRI-Ens attains the highest detection rate at matched false-detection
rates under our USA 30-kt standard and the published RITCF CMA 25.27-kt
standard. Every model within a panel shares the same sample, evaluation
label source, time window and metric definition; the training label sources
and reconstruction choices are described in Methods. WRPS1
\citep{horinouchi2025wrps1}, scored from its 73 published blocks on our
feature table, reaches $\mathrm{BSS_{dev}}=0.143$ on the shared test
sample ($n=2{,}892$) against 0.260 for CVRI-Ens (storm-cluster bootstrap
$\Delta\mathrm{BSS}=+0.117$, 90\% CI [$+0.045$, $+0.174$],
$p=0.010$). The rapid-intensification TC forecasting (RITCF)
contrastive-learning scheme \citep{wang2025ritcf}
was re-implemented on an independent pipeline using the published
description and released code, with the description taking precedence
where they conflict. The released implementation differs from the
described model and omits the data pipeline (Supplementary Note S4).
In this reconstruction (test 2020--2021, $n=1{,}057$), the rolling-pool
variant admits test-year RI labels, whereas the capped-pool variant excludes
them. The reconstruction nearly reproduces the reported POD (0.879--0.924
versus 0.910), but at much higher false-detection rates (46.5--54.9\% versus
9.4\%). No vote-threshold setting reproduces both reported values, and a
direct classifier on identical inputs attains higher ROC-AUC (0.854 versus
0.71--0.75). The published operating point would require an AUC of about
0.97 under an equal-variance binormal model (Supplementary Note S4).
The CMA-30 wind-source control re-scores the same frozen predictions on the
best-AUC curves (Supplementary Fig.~S4). CVRI-Ens retains the highest CSI
and the highest interpolated POD at 9.4\% FARate (from
\texttt{table\_figA\_best\_auc.csv}; 0.675 versus 0.567 for
the next-best entry). Its unrecalibrated probabilities have lower BSS than
WRPS1 under these labels (0.092 versus 0.134, from the same table).

\begin{figure}[!htbp]
\centering
\includegraphics[width=\textwidth]{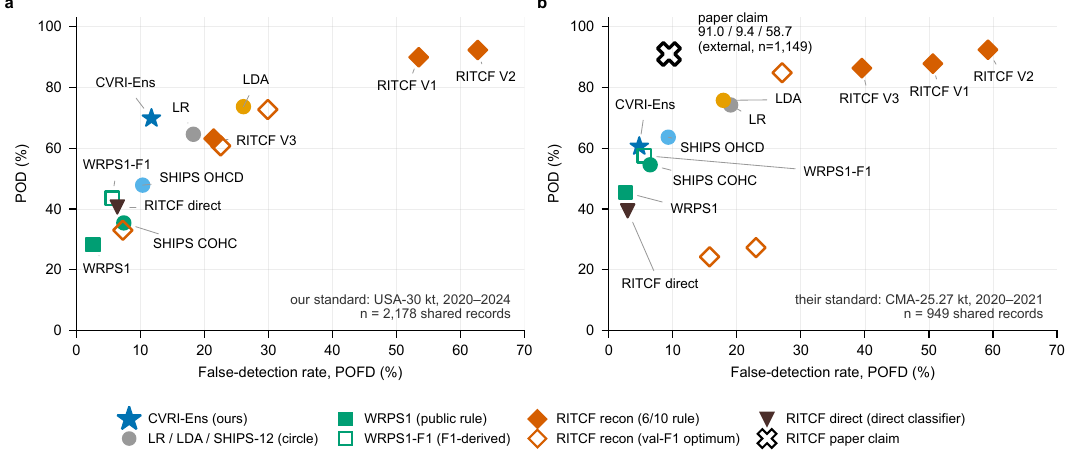}
\caption{\textbf{CVRI-Ens attains the highest detection rate at matched
false-detection rates under both label standards.} \textbf{a}, our
standard: USA-agency 30-kt rapid intensification (RI) labels, test
2020--2024, $n=2{,}178$ shared rows; \textbf{b}, the published
standard of \citet{wang2025ritcf}: China Meteorological
Administration (CMA) 25.27-kt labels, 2020--2021, $n=949$. Each point is
one model at its stated or inner-derived operating rule. The 9.4\% FARate
values quoted in the text are interpolated from the corresponding score-sweep
curves; the open marker
in \textbf{b} shows the operating point reported by
\citet{wang2025ritcf} on its own sample frame ($n=1{,}149$), which
is external to the shared sample. POD $=$ probability of detection;
FARate $=$ probability of false detection (POFD); FARatio $=$
$1-$precision. Panel-specific sample frames are listed in Supplementary Table S7.}
\label{fig:benchmarks}
\end{figure}

For comparison under our USA-agency 30-kt standard (2020--2024), we intersect
the 3,004 frozen holdout records with RITCF's CMA-wind/frame/storm eligibility
records and retain rows with no missing or infinite values in the 12 SHIPS predictors. 
This yields 2,178 records with 209 RI events (base rate 9.60\%, 
compared with 7.39\% for the full holdout; Supplementary Table S8 and Fig.~\ref{fig:benchmarks}a).
CVRI-Ens attains POD 0.699 at
an 11.7\% false-detection rate (FARate, equal to POFD). At the 9.4\%
false-detection rate reported for RITCF, it retains POD 0.670 against
0.517 for the next-best entry, interpolated from the primary score-sweep curves
(\texttt{iso\_pofd.csv}). On RITCF's own standard, using CMA 25.27-kt
labels for 2020--2021 ($n=949$;
Fig.~\ref{fig:benchmarks}b), CVRI-Ens ranks first in both ROC-AUC (0.896)
and PR-AUC (0.562; \texttt{table\_figB\_best\_loss.csv}) and retains POD 0.727
at that false-detection point (interpolated from the corresponding score-sweep curves; next best 0.636). 
The published operating point is shown for
reference (POD 0.910, FARate 9.4\%, FARatio 58.7\%,
$n=1{,}149$). WRPS1's development window extends through 2021, so its
scores include in-sample years in both panels and in the 2,892-row
comparison above.

% ============================================================ DISCUSSION
\section{Discussion}

This work couples observational causal discovery to deep prediction for
24-h RI warning. The discovered graph constrains training as an auditable
prior, and the system is evaluated under a common-standard protocol built
on proper scores. The graph links atmospheric and oceanic conditions to the
cyclone's dynamical and structural state, and through that state to 24-h
intensity change. Its eight sign-verified effects on \dv\ provide a compact,
physically interpretable prior.
Embedded in the model, the graph pathway yields a skill gain concentrated
near the estimation window, with the largest graph-removal effect in
LoRA-sf; control ablations link that gain to the discovered structure.
Under common evaluation standards, the ensemble outperforms
statistical--dynamical guidance and the reconstructed AI scheme in detection
at matched false-detection rates on both our label standard and the
published RITCF standard. To our knowledge, the
protocol-locked Top-5 ensemble's BSS of 0.259 is the highest reported for
probabilistic 24-h RI forecasts in the western North Pacific verified on
a strictly forward multi-year holdout against a fixed climatological
baseline. The closest prior values
\citep{eusebi2025salinity,horinouchi2025wrps1} come from cross-validated
protocols in which later years inform predictions of earlier ones, whereas
our evaluation preserves a strictly forward multi-year holdout.

Our causal estimates assume no unobserved confounding. The
synthetic-confounder refutation shifts every estimate with a non-empty
backdoor set, and the E-value analysis quantifies the confounding strength
needed to explain each effect away. The sensitivity configurations
probe lag selection, climate indices, latitude and OHC dependence, while
the DAG audit probes graph construction (Supplementary Table S1 and
Supplementary Notes S1--S2). The two wind sources assign the same RI label
to 95.4\% of the 2,178 panel-A records (Supplementary Table S8). The CMA-30 control preserves
the matched-FARate detection advantage and the highest CSI, while the
unrecalibrated probabilities yield lower BSS than WRPS1.

The framework is designed to be transferable: the prior can be rediscovered for other
basins or lead times and injected into larger backbones, and the
common-standard evaluation applies to any published scheme
with a stated operating rule. The discovery panel, frozen test
predictions and evaluation code are released with this paper, so that
every reported evaluation number can be reproduced.

% ============================================================ METHODS
\section{Methods}

\subsection{Study population and data sources}

The study population comprises all western North Pacific (WNP) tropical
cyclones (TCs) in the International Best Track Archive for Climate
Stewardship (IBTrACS, v04r01) \citep{knapp2010ibtracs,gahtan2024ibtracs} from 2000 to 2024,
with intensity, position and pressure taken from the USA-agency columns
(USA\_WIND, USA\_LAT, USA\_LON, USA\_PRES). The archive contains 725 WNP
systems in this period, of which 720 are label-eligible. The 17{,}643
label-eligible synoptic-time rows enter the population audit. A frame-availability rule
(\texttt{frame\_availability\_v1}) excludes 202 rows whose satellite imagery
is missing at the synoptic time (157 before the first available frame of the
storm, 30 after the last, 15 across an internal frame gap), with the reason
recorded per row; 17{,}441 rows remain. Rows are split by storm year into a
fit period (2000--2017; 12{,}619 rows, 522 systems), an inner period
(2018--2019; 1{,}818 rows, 66 systems) used for threshold selection, early
stopping and ensemble selection, and a test period (2020--2024; 3{,}004 rows,
131 systems) evaluated exactly once after the final protocol lock. One 2017
storm in the system-level eligibility census is absent from the three analysis
periods because its 12 rows have 24-h target times in 2018, giving 719 systems
overall.

Atmospheric predictors come from the ERA5 reanalysis
\citep{hersbach2020era5} on pressure levels 1000--200~hPa and single-level
fields, sampled on storm-centred inner-core (0--2$^\circ$) and outer-area
(2--8$^\circ$) annuli. Ocean heat content (OHC) above the 26$^\circ$C
isotherm comes from the GLORYS12V1 ocean reanalysis
\citep{lellouche2021glorys}, taken from completed daily windows
(backward selection to the six-hourly grid). Maximum
potential intensity is computed with tcpyPI \citep{gilford2021tcpypi} and
enters the baseline predictor table (Supplementary Table S9).
Satellite infrared imagery comes from the HURSAT-B1 v07b archive
\citep{knapp2011hursat}: storm-centred black-body brightness temperature
(TBB) frames at 224$\times$224 px, up to five frames per record
($t-24$\,h to $t$ at 6-h steps). Distance to land is computed from the
Global Self-consistent, Hierarchical, High-resolution Geography (GSHHG)
coastline \citep{wessel1996gshhg}. Table~\ref{tab:variables} lists the
environmental variables and their physical definitions.

\begin{table}[!htbp]
\centering
\caption{Environmental variables and physical definitions. Inner core means
0--2$^\circ$ and outer area means 2--8$^\circ$ around the storm centre.}
\label{tab:variables}
\scriptsize
\setlength{\tabcolsep}{3.5pt}
\resizebox{\textwidth}{!}{%
\begin{tabular}{llll}
\toprule
Variable & Meaning & Construction & Source \\
\midrule
\multicolumn{4}{l}{\emph{Best-track state}} \\
VMAX & Current intensity & USA 1-min sustained wind (kt) & IBTrACS \\
PER & Recent intensification & VMAX($t$)$-$VMAX($t-6$h) & IBTrACS \\
SPD & Translation speed & Great-circle displacement rate between best-track positions (kt; nmi h$^{-1}$) & IBTrACS \\
LAT & Latitude & Storm-centre latitude & IBTrACS \\
LAND\_DIST & Distance to land & Great-circle distance (km) & GSHHG coastline \\
\multicolumn{4}{l}{\emph{ERA5 atmospheric fields}} \\
SHRD & Deep-layer shear & 850--200-hPa wind difference, outer area & ERA5 \\
SHEAR\_VMAX\_ANGLE & Shear--motion angle & Shear direction relative to TC motion, 0--180$^\circ$ & ERA5 + best track \\
D200 & Upper-level divergence & 200-hPa divergence, outer area & ERA5 \\
RHLO & Low-level humidity & 850--700-hPa mean relative humidity, outer area & ERA5 \\
RHHI & Mid/upper-level humidity & 500--300-hPa mean relative humidity, outer area & ERA5 \\
VORT & Low-level vorticity & 850-hPa relative vorticity, inner core & ERA5 \\
PV & Potential vorticity & 500-hPa potential vorticity, inner core & ERA5 \\
SST\_OA & Outer-area SST & Sea-surface temperature, outer area & ERA5 \\
SST\_IC & Inner-core SST & Sea-surface temperature, inner core & ERA5 \\
WARM\_CORE & Warm-core anomaly & 200-hPa temperature anomaly, inner minus outer & ERA5 \\
MFC & Moisture flux convergence & 850-hPa moisture flux convergence, outer area & ERA5 \\
SLHF\_IC & Inner-core latent heat flux & Surface latent heat flux, ocean to atmosphere & ERA5 \\
DRY\_AIR & Dry-air intrusion & Outer-area fraction with 700--500-hPa relative humidity $<40$\% & ERA5 \\
W500 & Mid-level pressure velocity & 500-hPa pressure velocity, outer area; negative is ascent & ERA5 \\
\multicolumn{4}{l}{\emph{GLORYS12V1 ocean fields}} \\
OHC\_IC & Inner-core OHC & OHC above 26$^\circ$C isotherm, inner core & GLORYS12V1 \\
OHC\_OA & Outer-area OHC & OHC above 26$^\circ$C isotherm, outer area & GLORYS12V1 \\
\bottomrule
\end{tabular}}
\end{table}

\subsection{Outcomes and labels}

The continuous outcome is the 24-h intensity change
\begin{equation}
\dv=\operatorname{clip}\left[\mathrm{VMAX}(t+24\,\mathrm{h})-\mathrm{VMAX}(t),-55,55\right]\ \mathrm{kt},
\end{equation}
where the limits are the 1st and 99th percentiles fitted on fit-period data.
Rapid intensification (RI) is computed from the unmodified difference,
$\mathrm{RI}=\mathbf{1}\{\mathrm{VMAX}(t+24\,\mathrm{h})-\mathrm{VMAX}(t)\ge30\ \mathrm{kt}\}$,
following the standard 30-kt-per-24-h definition \citep{kaplan2010rii}.
The label time is the nearest best-track entry within $\pm6$~h of $t+24$~h
(130 test rows therefore deviate from the exact 24-h horizon: 127 at
18~h and 3 at 21~h).
The RI base rate is 7.39\% in the test period and 7.91\% over the
2000--2019 development period; the latter is fixed as the climatological
reference for all skill scores below. Environmental variables are z-scored
with scalers fitted on the fit period only. The Yeo--Johnson transform \citep{yeo2000transform} of
\dv\ is also fitted on the fit period only, and is used in discovery and as
the auxiliary regression target.

\subsection{Masked causal discovery}

Causal discovery uses PCMCI+ \citep{runge2020pcmcip} with partial-correlation
(ParCorr) conditional-independence tests, maximum lag $\tau_{\max}=4$
(24~h at 6-h resolution) and $\mathrm{pc\_alpha}=0.01$, run on all
label-eligible fit-period rows (12{,}647 rows from 522 systems, feature
times up to 25 December 2017 18:00 UTC). Discovery does not require
satellite frames and therefore retains the 28 fit-period rows later
excluded by the frame-availability rule; 12{,}619 of these rows enter the
modelling panel. Storms are independent
realizations, so no lagged link crosses storm boundaries. The panel is
aligned on a complete six-hourly grid; padding and long-gap-filled cells
are masked, and samples masked in the treatment, outcome, or any
conditioning variable are excluded from each test
(\texttt{mask\_type=xyz}). Short gaps of at most three steps are filled
by linear interpolation so that lagged indices stay aligned within a
storm, but every filled cell remains masked and is excluded from the
affected tests. The target is never interpolated.

Skeleton selection uses unadjusted test p-values. Benjamini--Hochberg (BH)
correction is applied to lagged links, while contemporaneous links retain
unadjusted p-values \citep{benjamini1995fdr}. Meteorological priors prohibit
the target from causing any predictor, constrain derived-variable directions
and lags, and allow a limited set of contemporaneous links with fixed
orientation into the target. The resulting graph contains 81 lagged links at
$q\le0.05$ and six contemporaneous links at unadjusted $p\le0.05$. The ten
further links above these thresholds are excluded from the
model's graph branch, while the identification DAG below is built from the
full discovery output so that adjustment sets account for them. Conditional-independence-based discovery identifies graphs up to
their Markov equivalence class \citep{pearl2009causality,runge2019pcmci}; we therefore
treat the graph as a set of candidate links and subject effect signs to the
identification and refutation pipeline below before any use in training.

\subsection{Identification and effect estimation}

For effect estimation, directed links are collapsed across lags by retaining
the largest absolute coefficient for each ordered variable pair, and an
identification directed acyclic graph (DAG) is built from the collapsed
graph with greedy weakest-edge cycle breaking and a small set of structural
edges inserted as modelling choices (recorded in the DAG audit, Supplementary Note S1).
DoWhy \citep{sharma2020dowhy} returns backdoor adjustment sets from this DAG.
Effect estimates are computed per one-standard-deviation increase of each
treatment: by ordinary least squares for \dv\ (reported in kt per 24~h per
SD) and by a binomial generalized linear model for RI (reported as a
conditional log-odds coefficient per SD). A treatment with no directed
path to the outcome has no identified total effect, and its regression
output is reported as an association. D200 has no directed path to \dv\
in the identification DAG, and its RI estimate rests solely on the inserted
treatment$\to$RI edge, so both values are reported as associations. The
protocol-locked rule therefore excludes D200 from the trusted set. The
strict-lag and no-OHC configurations find directed paths for D200 and are
reported as sensitivity analyses (Supplementary Note S2).

Uncertainty uses 200 block-bootstrap resamples of whole storms (seed 42),
with BH q-values computed separately for the \dv\ and RI families. We report
95\% intervals for these effect estimates, conditional on the locked
identification DAG; skill contrasts on the much smaller holdout are instead
tested at the locked significance level $\alpha=0.10$ with
90\% intervals (see below).
Robustness is probed with four refuters (placebo treatment, 90\% subset,
synthetic unobserved confounder, offshore subset) and approximate E-values
at the estimate and at the nearer-to-null confidence limit
\citep{vanderweele2017evalue}. A
treatment sign enters the model's constraint set (the \emph{trusted signs},
defined on the \dv\ scale) only if the bootstrap $p<0.05$, at least two
refuters pass, and a directed
identification path to \dv\ exists. Eight treatments qualify: SHRD$-$, RHLO$+$,
VORT$-$, PV$-$, WARM\_CORE$-$, W500$-$, OHC\_IC$+$, OHC\_OA$+$. Sensitivity
configurations (strict lag-$\geq$1 discovery, Madden--Julian oscillation
and Ni\~no-3.4 indices
added to adjustment sets, latitude added to the OHC\_OA adjustment set, and
a no-OHC graph) are reported in Supplementary Note S2.

\subsection{The causal prior as model inputs}

The outputs of the discovery pipeline enter the model through three channels.
The 87 retained graph edges are passed to a graph neural network (GNN) with
edge attributes (lag/$\tau_{\max}$, sign of the coefficient, absolute
coefficient). The eight trusted signs anchor a sign--gradient loss and a
counterfactual term over the same treatments. The normalized absolute edge
coefficients anchor the adjacency loss (below). Every training run logs
cryptographic SHA-256 checksums of the graph and ATE files, so the prior
version is auditable per checkpoint.

\subsection{Model architecture}

CausalVisionRI fuses three input sources (Fig.~\ref{fig:architecture}). Satellite TBB sequences
are encoded by a vision transformer \citep{dosovitskiy2021vit}, with two initialization
routes across the arm pool: the LingBot-Vision base student (ViT-B/16)
\citep{fu2026lingbotvision} as an external initialization, and
a from-scratch ViT-S/16 pre-trained in-domain as a masked autoencoder
(MAE; below) \citep{he2022mae}. Fine-tuning modes span frozen, low-rank
adaptation (LoRA, rank 16) \citep{hu2022lora}, partial-block and full fine-tuning, combined
with temporal-window variants (single-frame, two-frame, three-frame and
five-frame inputs); all five ensemble members use the LingBot route. The 21
environmental variables over five history
steps are encoded by a multilayer perceptron (hidden width 256). Their
per-record histories also supply node features to a two-layer GNN (hidden
width 128, four attention heads, edge dimension 3). The GNN has 21
environmental nodes and one learnable \dv\ readout node with no input
features. Message passing is restricted to the 87 prior edges, and the
causal token concatenates the readout-node state with the mean environmental
node state. Fusion is a gated cross-attention block \citep{vaswani2017attention}
($d_{\mathrm{model}}=256$, four heads): the pooled visual token queries the
environmental and causal tokens, and a per-row sigmoid gate $g$ scales the
attention output before the residual connection, so the fused attention
output can be down-weighted row by row. A linear head outputs the 24-h RI
probability, trained with a focal loss \citep{lin2017focal} ($\gamma=2$, $\alpha=0.25$), and an
auxiliary linear head predicts Yeo--Johnson-transformed \dv.

The causal prior is used to constrain the model through the following losses.
They include an adjacency cross-entropy
matching each target node's incoming-edge attention to its
coefficient-normalized prior weights (weight 1.0), a sign--gradient term
penalizing the batch- and history-mean input gradients whose signs oppose
the trusted effects (weight 2.0), and a counterfactual term penalizing \dv\
predictions that increase when each trusted treatment is shifted by one
standard deviation against its effect sign at all five history steps
(weight 0.02). The input gradient is a local surrogate for the sign of the
estimated average treatment effect. Both response terms act on the fused
model's transformed \dv\ output. Causal losses are scaled by 0.01 during
fine-tuning.

\subsection{Satellite pre-training}

For the from-scratch route, a ViT-S/16 backbone is pre-trained as an MAE
on storm-centred frames drawn from a
multi-satellite raw corpus of 304{,}518 frames spanning all basins, via
two routes: a WNP route restricted to the study population's systems and
a global all-basin route. The MAE settings are a mask ratio of 0.75,
normalised-pixel loss, at most 400 epochs with early-stopping
patience 20, batch size 256, base learning rate $1.5\times10^{-4}$, weight
decay 0.05 and a 40-epoch warmup. On each route, fit-period frames are used for
training and inner-period frames for validation; test years are excluded.
Same-time frames from different satellites are retained as sensor
augmentation; only non-physical fill pixels are masked. The WNP route
reached its best validation loss at epoch 96 and the global route at
epoch 75. Downstream, frames are restricted
to the quality-controlled set of 153{,}295 unique (storm, time) frames so
that every panel row aligns with exactly one image sequence.

\subsection{Training, selection and ablation protocol}

Downstream models are trained in two stages: full-network \dv-regression
pre-training (30 epochs, learning rate $10^{-3}$) followed by RI fine-tuning
(40 epochs, learning rate $3\times10^{-4}$). The backbone learning rate is
scaled by 0.1 and weight decay is 0.01. Both stages use early stopping with
patience 10, on inner-period \dv\ mean squared error in the first stage and
inner-period RI skill in the second. All 22 development arms were trained on
fit-period data and selected on inner-period data only: arms are ranked by
per-year inner Brier skill score (BSS), evaluated against each year's own
climatology, with worst-year-first ordering. Family champions receive a
protected slot only when they rank within the
overall top eight, and remaining slots are filled by overall rank. The top five members are
averaged with equal weights to form CausalVisionRI-Ensemble (CVRI-Ens):
lora16\_sf, lora16\_temp5, temp2, partial4\_sf, temp5\_slim, referred to
below by the short names LoRA-sf, LoRA-T5, T2, Part4 and T5 respectively
(full run ids in Supplementary Table S2). The
evaluation threshold
(0.225 for CVRI-Ens) maximizes the inner-period F1 score (harmonic mean
of precision and recall) and is fixed
before any test evaluation. After the final protocol lock (entity list,
thresholds, comparison families and the protocol-locked primary contrast
\texttt{C2\_vs\_M0\_BSS\_dev}), test-period predictions were computed once
and frozen; a signed receipt records completion (lock files and audit
trail in Supplementary Note S5).

The ablation panel comprises 40 entities (five members $\times$ eight
conditions) trained under the same protocol with shared member
initializations and fixed shuffle generators, each condition evaluated
against its member's full causal configuration (M0). The conditions are:
C1 removes all causal losses
(GNN retained); C2 removes the GNN branch (causal losses inactive); C3s1--3
replace the prior graph with shuffles that preserve each node's degree in
the environmental subgraph, under three
seeds (same GNN capacity, same losses); C4adj, C4cf and C4sign retain only
the adjacency, counterfactual, or sign--gradient loss. Because C3 keeps the
input dimensionality and loss weights unchanged, it
probes the contribution of this specific graph.

\subsection{Statistical--dynamical and AI baselines}

Four reference systems are evaluated under the identical protocol
(same rows, same threshold rule, same metrics): a 12-predictor SHIPS-style
linear discriminant analysis (LDA) reconstruction in two oceanic variants
(climatological OHC, COHC; time-dependent daily OHC (OHCD)) \citep{kaplan2010rii,
demaria2005ships} (the twelve predictors, including the tcpyPI-based
potential-intensity term, are listed in Supplementary Table S9), a
21-predictor panel LDA, and class-balanced logistic regression (LR), all
fitted on fit-period rows with fit-median imputation of missing predictors.
LR is retained as an uncalibrated reference. Two published
schemes are additionally evaluated under common standards
(Section~\ref{sec:eval}): the WRPS1 knowledge-block scheme, reconstructed
from the 73 published blocks (18 positive, 55 negative) with the published
offset and probability mapping and scored directly on our SHIPS-12 feature
table with the stated unit conventions; and the RITCF contrastive scheme
\citep{wang2025ritcf}, re-implemented on an independent data pipeline
using the published description and released code, with the description
taking precedence where they conflict. All undocumented design choices are
recorded as deviations in Supplementary Notes S3 and S4.

\subsection{Evaluation and statistics}
\label{sec:eval}

The primary metric is the Brier skill score against the fixed development
climatology, $\mathrm{BSS_{dev}} = 1 - \mathrm{BS}/\mathrm{BS_{clim}}$, with
$\mathrm{BS_{clim}} = \pi^2 + (1-2\pi)\bar{y}$: $\pi$ is the 2000--2019 RI
rate of 0.0791 held fixed for every model, and $\bar{y}$ is the observed
frequency of the evaluated sample. Discrimination is summarized by the
area under the receiver-operating-characteristic curve (ROC-AUC) and the
precision--recall curve (PR-AUC) \citep{saito2015pr}; thresholded performance uses the
probability of detection (POD), probability of false detection (POFD),
false-alarm ratio (FARatio $=1-$precision) and critical success index (CSI)
at the inner-derived threshold. Uncertainty uses storm-cluster paired
bootstrap resampling (2{,}000 resamples of whole storms, seed 42) of
score differences, with Holm adjustment \citep{holm1979procedure} inside locked comparison
families; significance is reported at $\alpha=0.10$ with 90\% confidence
intervals (CIs), and every contrast reported as significant on the full
test period also passes at $\alpha=0.05$. Common-standard comparisons evaluate every model on the
intersection of eligible rows under a single evaluation label source, time
window and metric definition, with published operating rules executed as
stated (WRPS1 $p_u=0.5$ mapping; RITCF 6-of-10 vote). CVRI-Ens, SHIPS-12,
LDA, LR and WRPS1 use USA-side training or development labels, whereas
RITCF uses CMA labels. The USA-30 panel therefore places RITCF in a
cross-label evaluation, and the CMA-25.27 panel places the USA-developed
models in a cross-label evaluation. Two fusion
diagnostics are reported: the mean activation of the gate that scales the
joint environmental and causal attention output (gate mean), and the
head-averaged share of cross-attention assigned to causal tokens (causal
share; the no-preference reference is 0.5 for two context tokens).

% ============================================================ AVAILABILITY
\section{Data availability}
The processed panel with the fit/inner/test split markers and row-level
keys, the discovered causal graph, effect-estimate tables and trusted-sign
file (with cryptographic SHA-256 manifests), the frozen test-period predictions of all
evaluated entities, and the satellite-frame index are released with
this paper in the project data repository; ERA5, GLORYS12V1 and HURSAT-B1 are
available from their respective providers (European Centre for
Medium-Range Weather Forecasts (ECMWF) Climate Data Store (CDS),
Copernicus Marine, National Centers for Environmental Information (NCEI)),
and IBTrACS from the National Oceanic and Atmospheric Administration (NOAA) NCEI.
The project data record is available at
\url{https://doi.org/10.5281/zenodo.23768442}.

\section{Code availability}
All pipeline code, including causal discovery, prior packaging,
pre-training, training, ablation, evaluation and figure generation, is
released with this paper under an open licence (MIT License),
including the protocol-lock files that fix the evaluation protocol.
The code is available at \url{https://github.com/Clrs17/CausalVisionRI}.

\bibliographystyle{plainnat}
\bibliography{references}

\section*{Author contributions}
X.L.\ and Y.J.\ designed the study, built the pipeline and ran the
experiments. Y.J.\ built the dataset. X.L.\ and Y.D.\ analysed the results.
X.L.\ and Y.D.\ drafted the manuscript. S.S.\ supervised the work and
revised the manuscript.

\section*{Competing interests}
The authors declare no competing interests.

\section*{Acknowledgements}
Supported by the National Natural Science Foundation of China (Grant No. U2442203), 
and Project 202610335008 of the 
National Training Program of Innovation and Entrepreneurship for Undergraduates.  
The authors acknowledge the Copernicus Climate Change Service (C3S) for providing 
the ERA5 reanalysis, and the Copernicus Marine Environment Monitoring Service (CMEMS) 
for providing the GLORYS12V1 ocean reanalysis. The authors are grateful to Zhiyang Ye 
(School of Remote Sensing and Information Engineering, Wuhan University) for helpful 
discussions on remote sensing imagery processing.
\end{document}

\maketitle

\noindent\emph{Abbreviations used in this supplement.} ATE, average
treatment effect; AUC, area under the receiver-operating-characteristic
curve; BSS, Brier skill score; CI, confidence interval; CMA, China
Meteorological Administration; CVRI-Ens, CausalVisionRI-Ensemble; DAG,
directed acyclic graph; ENSO, El Ni\~no--Southern Oscillation; FDR, false
discovery rate; GNN, graph neural network; MJO, Madden--Julian oscillation;
OHC, ocean heat content; PCMCI+, a conditional-independence
causal-discovery algorithm \citep{runge2020pcmcip}; RI, rapid
intensification; SD, standard deviation.

\section*{Supplementary Notes}

\subsection*{Supplementary Note S1: Causal-discovery details}

The identification DAG is built from the full discovery output (97
directed links, including the ten with $q>0.05$, so that adjustment sets
account for them), collapsed to 77 directed variable pairs by retaining
the largest absolute coefficient across lags per ordered pair. The DAG
inserts 12 structural edges (the target$\to$RI edge defining the binary
estimand, ten treatment$\to$RI edges, and W500$\to$target) and removes 26
edges by greedy weakest-edge cycle breaking; one further candidate
(SST\_OA$\to$W500) was skipped because it would recreate a cycle, and
seven contemporaneous adjacencies remain unorientable and are excluded from
the DAG. Deleting the inserted W500$\to$target edge leaves W500's identified
effect unchanged. Across the admissible identification DAG constructions in
the released audit, the OHC\_OA effect ranges from $+3.39$ to $+8.92$, and
every trusted sign remains stable. All construction
choices are enumerated in the released DAG-audit files
(\texttt{dag\_audit\_all/}), and the per-treatment identification status
is given in \texttt{identification\_status.csv}. Effect signs are used in
training only for treatments passing 200 storm block bootstraps
($p<0.05$), at least two of four refuters, and the directed-path check.

On the \dv{} outcome, all ten treatments pass the placebo and 90\%-subset
refutations. The offshore-subset refuter fails for RHLO, OHC\_IC and
OHC\_OA, and the synthetic unobserved-confounder refutation shifts estimates
for all treatments with a non-empty backdoor set; Supplementary Table S1
therefore reports E-values (1.13--2.08 on the \dv{} scale), i.e.\ the
minimum confounding strength needed to explain away each estimate. For
\dv, $d=|\mathrm{effect}|/20.1321$ and $RR\approx\exp(0.91d)$, where
20.1321~kt is the SD of \dv\ and 0.91 is the log-odds conversion factor. For
RI, $OR=\exp(\beta)$ is converted using $RR=OR/(1-p_0+p_0OR)$ with
$p_0=1014/12647$, inverting $RR$ for protective effects. The
assumptions behind the discovery, how each is handled here and what follows
if one of them is violated are collected in Supplementary
Table~\ref{tab:assumptions}.

\subsection*{Supplementary Note S2: Sensitivity configurations}

The four sensitivity configurations use the same fit-period panel
(2000--2017; 12{,}647 rows), estimator, lag limit and lagged-link FDR.
The link, variable and adjustment-set changes are specified below.

\textbf{Strict lag-$\geq$1 discovery} (no non-self contemporaneous target
links) yields 91 significant links and the same qualitative target
structure: the direct lag-1 OHC\_OA link into \dv\ persists, joined by
PER (lag 1), VMAX (lag 2) and D200 (lag 2). The strict DAG also changes
the backdoor adjustment sets, so effects are re-estimated. The trusted \dv\ effects are essentially unchanged (OHC\_OA $+7.07$
versus $+6.98$ at baseline; every other trusted-sign estimate retains its
sign and remains within a few percent of its baseline value).

\textbf{Climate-index adjustment} appends the MJO index
(OMI1/OMI2) \citep{kiladis2014omi} and Ni\~no-3.4 indices to every baseline
adjustment set. No estimate changes sign; \dv-scale magnitudes shift by at
most 4.4\% (OHC\_OA $+6.98 \to +7.03$). These indices vary mostly between
storms, so this configuration probes season-scale confounding instead of
within-storm pathways.

\textbf{Latitude adjustment} adds LAT to the OHC\_OA adjustment set
(correlation between LAT and OHC\_OA $=-0.63$): the effect remains large
and significant ($+6.48$ [$+5.52$, $+7.31$] versus $+6.98$ [$+6.06$,
$+7.82$] at baseline; RI outcome $+0.67 \to +0.76$), so the large
outer-area OHC effect is not explained by latitude.

\textbf{No-OHC configuration} removes both OHC variables from graph
variables, treatments and effect analyses (75 significant links; target
parents reduce to VMAX links). On the \dv\ scale, the non-OHC estimates
retain their signs (PV $-4.54$, WARM\_CORE $-5.59$, W500 $-3.95$, SHRD
$-4.79$ unchanged), and D200 acquires a directed path with an adjusted
estimate of $+4.99$.

All variant outputs (graphs, effect tables, manifests with input hashes)
are released with the code package.

\subsection*{Supplementary Note S3: WRPS1 reconstruction}

WRPS1 \citep{horinouchi2025wrps1} was evaluated by scoring its 73
published knowledge blocks (18 positive, 55 negative) directly on our
SHIPS-12 feature table, using the published offset $c=-0.3536$,
coefficient $\alpha=0.275$ and probability mapping
$P(p_l)=1.134\,p_l-0.296\,p_l^2$, with the published unit conventions
(shear in 0.1~kt, temperatures in 0.1$^\circ$C). Two typographical
irregularities in the published supplement were handled as recorded:
printed block row 55 duplicates row 1 exactly and was counted once, and
block number 36 appears twice with two distinct conditions, both kept.
All 73 block conditions and weights were checked line by line against the
published supplement images. The evaluation subset here spans all periods
and comprises the 16{,}718 rows with all twelve factors finite (the test
subset used in the main text is the 2,892-row intersection); the per-factor distributional
alignment statistics (factor means and RI minus non-RI directions, all
consistent with the block logic) are released with the comparison
artifacts.

\subsection*{Supplementary Note S4: RITCF reconstruction}

We re-implemented the RITCF contrastive scheme
\citep{wang2025ritcf} on an independent data pipeline (HURSAT-B1 frames,
ERA5 fields, CMA best-track labels), following the published description
and the authors' released code. Two points in the released code
contradict the published description (the 32-dimensional embedding and
the normalization ranges of its Eq.~1), and there we follow the paper.
For branch weight sharing and the pair-level threshold, which are unspecified
in the paper, we use shared weights and the inner-period F1 optimum. Every such choice is recorded in the released
deviation log. Variants: V1 (faithful, with a rolling reference pool that
admits within-test-year RI labels, following the authors' protocol), V2 (pool capped at 2019), V3 (the released
repository skeleton: unshared weights, scalar distance), V4 (no voting)
and V4direct (a direct classifier on identical inputs).

\textbf{Training behaviour.} Under a fixed 60-epoch budget, every
siamese variant peaks at validation AUC within the first one to two
epochs and declines overall afterwards, while the training pair
loss approaches zero; pair distances of positive and negative pairs
overlap almost completely. The direct classifier on identical inputs
follows the same early-peak, late-decline pattern, so the early peak is
not specific to the contrastive objective. Checkpoint selection is reported under three
criteria (best validation loss, best validation AUC, last epoch) in
full; for most variants the loss- and AUC-selected checkpoints coincide,
and the last-epoch criterion is reported alongside. The primary cross-model
comparison (main-text Fig.~5) uses best-loss checkpoints; Supplementary
Figs.~S3 and S4 use best-AUC score sweeps for comparability. On the 25.27-kt
protocol, the val-F1 operating points for V1--V4 are taken from the operating
table associated with that comparison because the best-AUC table lacks these variants; V4direct
uses its best-AUC score sweep and operating table.

\textbf{Operating points (best-AUC criterion).} On the published protocol
(CMA 25.27-kt labels, test 2020--2021, $n=1{,}057$): V1 87.9\% POD at
46.5\% FARate; V2 92.4\%/54.9\%; V3 90.9\%/51.8\%; V4direct AUC 0.854.
On the CMA 30-kt protocol (test 2020--2024, $n=2{,}385$): V1 92.5\%/51.7\%;
V2 95.8\%/60.8\%; V3 70.8\%/20.9\%. Across the full vote-threshold sweep
(1/10 to 10/10) no operating point approaches the reported combination
(91.0\% POD at 9.4\% FARate); attaining POD 0.91 at FARate 0.094 would
require a sample-level AUC of about 0.97 under an equal-variance binormal
model, against 0.71--0.75 measured for the reconstructed siamese family and
0.854 for the direct classifier on that sample. The rolling reference
pool in V1 lowers POD by 4.5 pp and FARate by 8.4 pp relative to V2
(V1 POD 0.879, FARate 0.465; V2 0.924, 0.549), but does not by itself
account for the discrepancy.

\subsection*{Supplementary Note S5: Evaluation protocol and audit trail}

The protocol is fixed by \texttt{selection.json}, which records the 22
arms and the selection rule, and \texttt{final.json}, which records the 49
evaluated entities and four reference systems.

The primary comparison definition is in \texttt{comparison\_secondary.json};
the best-AUC definition is in
\texttt{comparison\_secondary\_best\_auc.json}. Both carry row-key hashes,
thresholds and metric sets. \texttt{test\_receipt.json} records the completed
once-only test evaluation.
The frame-availability rule excluded 202 rows (157 before a storm's first
available satellite frame, 30 after the last, 15 across an internal gap),
each with a recorded reason. All 22 arms and 40 ablation entities carry
completion markers with artifact hashes; the released audit record includes
the rerun manifest.
Pre-training endpoints: WNP route best at epoch 96, global route at epoch
75 (validation losses 0.7777 and 0.7629).

\section*{Supplementary Tables}
\input{sm_tables}

\begin{table}[!htbp]
\centering
\caption{\textbf{Discovery assumptions, their handling, and the
consequences if violated.}}
\label{tab:assumptions}
\scriptsize
\begin{tabular}{p{3.2cm}p{5.6cm}p{5.6cm}}
\toprule
Assumption & Handling here & If violated \\
\midrule
Causal sufficiency (no unobserved confounder) & All candidate drivers
drawn from reanalysis and best-track fields; E-values quantify the
confounding strength needed to explain each effect away (Table S1) & Estimates may be biased in either direction;
E-values give the minimum confounding strength required to explain an
effect away (DV24: 1.13--2.08; RI: 1.33--3.39) \\
Stationarity of links & Discovery on the 2000--2017 fit window only;
four sensitivity configurations re-estimated (Note S2) & Within-window
sensitivity does not test stationarity; the configurations probe lag
selection, climate indices, latitude and OHC dependence (Note S2), and
alternative DAG constructions are audited separately (Note S1) \\
Faithfulness (observed independencies reflect the graph) & Masked tests
with \texttt{mask\_type=xyz}; BH control at $q\le0.05$ for lagged links and
unadjusted $p\le0.05$ for contemporaneous links & Weak-signal
links may be missed (e.g., MFC), limiting detection power rather than
demonstrating unfaithfulness \\
Temporal ordering & Six-hourly storm-aligned grid; target never
interpolated & Lagged-link orientations rely on the six-hourly ordering;
lag-0 links are handled separately (flagged or removed in the strict
configuration) \\
\bottomrule
\end{tabular}
\end{table}

\clearpage
\section*{Supplementary Figures}

\begin{figure}[!htbp]
\centering
\includegraphics[width=\textwidth]{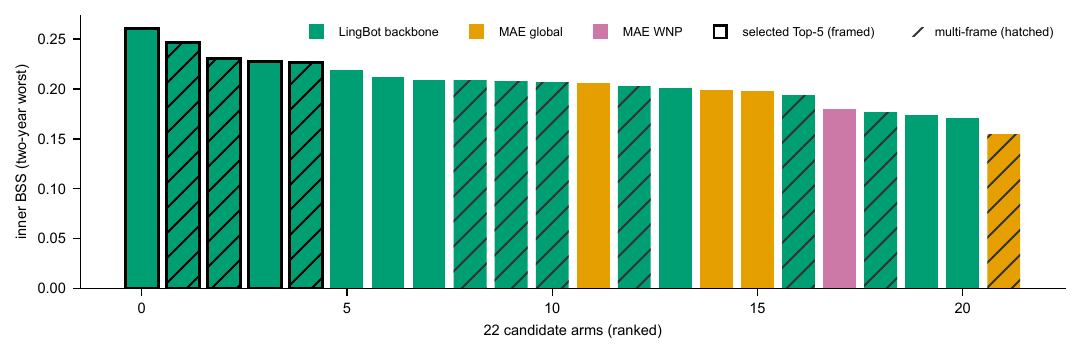}
\caption{\textbf{All 22 development arms, ranked by two-year-worst
inner-period (2018--2019) Brier skill score (BSS); the framed bars form
the selected Top-5 equal-weight ensemble.} Colour denotes the backbone
origin (LingBot-Vision initialised, or masked-autoencoder routes on the
global or western North Pacific corpora); hatching denotes multi-frame
temporal variants.}
\label{fig:s_selection}
\end{figure}

\begin{figure}[!htbp]
\centering
\includegraphics[width=0.7\textwidth]{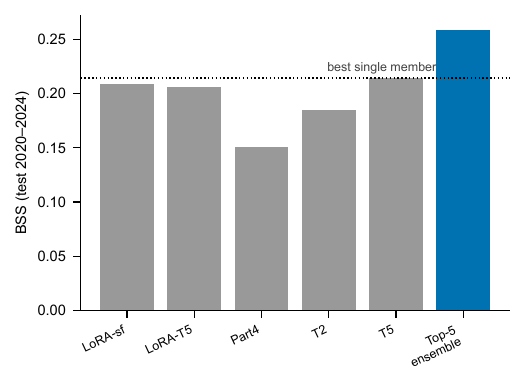}
\caption{\textbf{The equal-weight Top-5 ensemble exceeds every member on
the test period (2020--2024).} Bars: test BSS of the five members and of
the ensemble (dotted line: best single member).}
\label{fig:s_ensemble}
\end{figure}

\begin{figure}[!htbp]
\centering
\includegraphics[width=\textwidth]{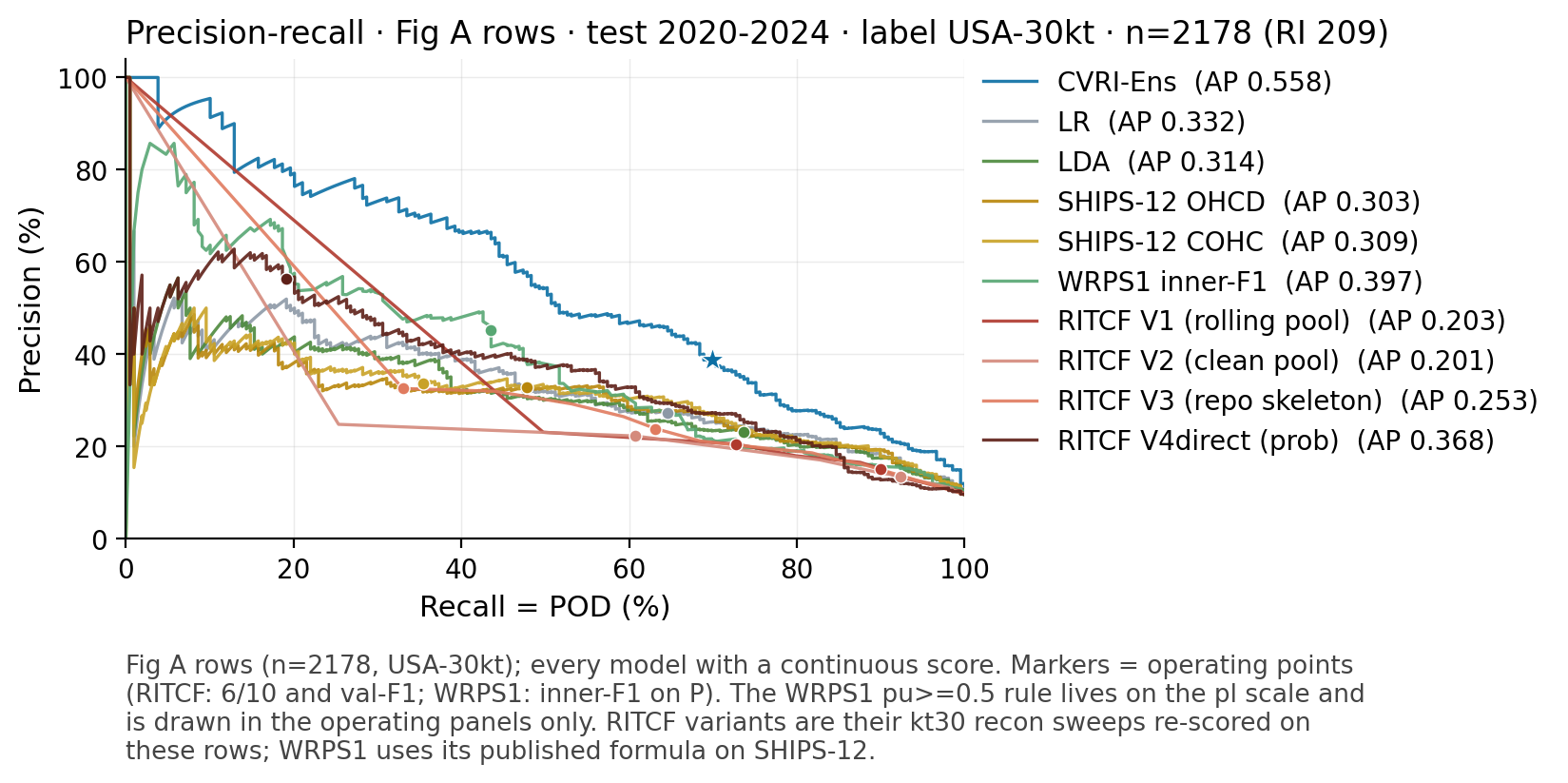}
\includegraphics[width=\textwidth]{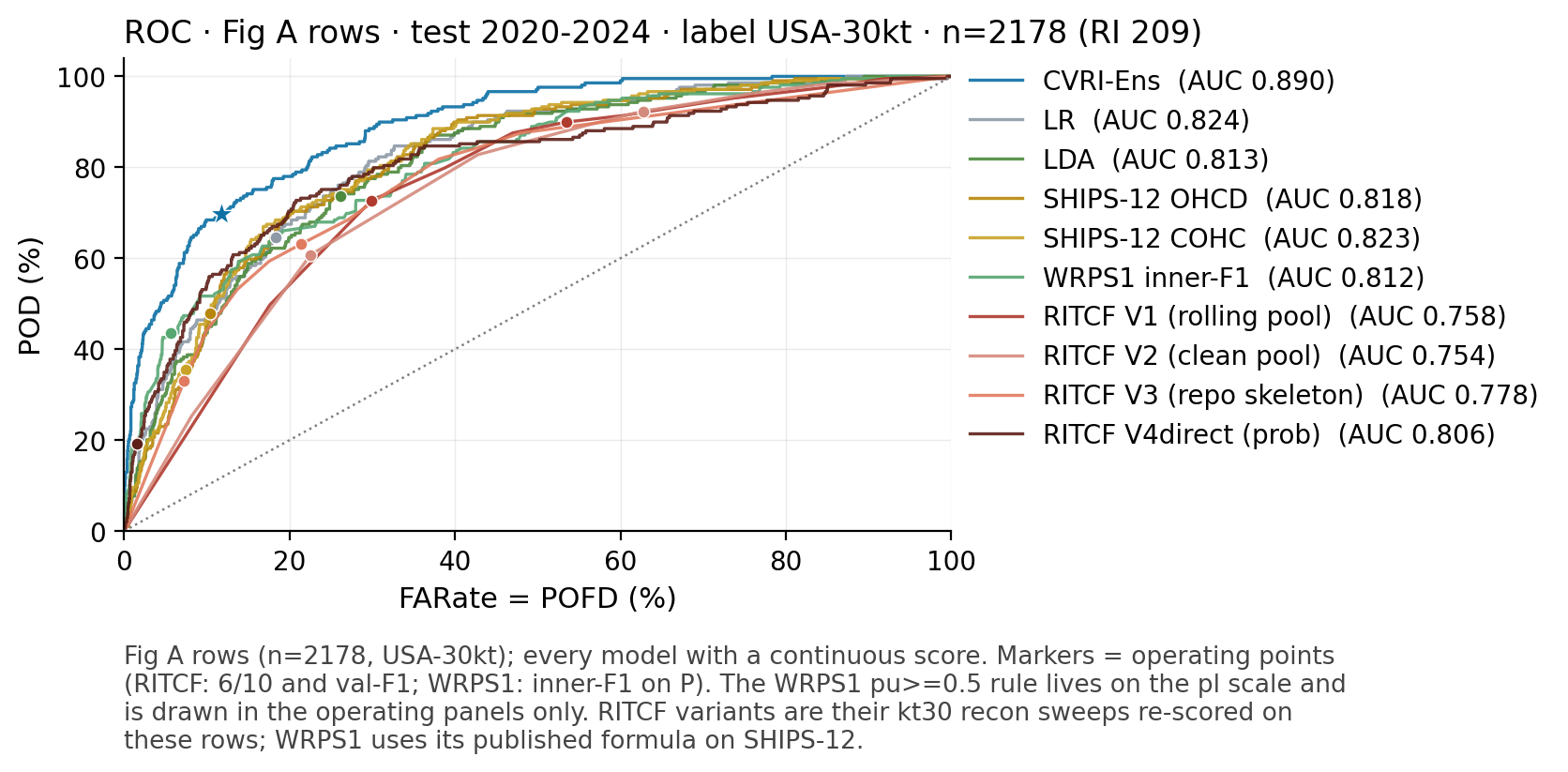}
\caption{\textbf{CVRI-Ens lies at or above the reference precision--recall curves
for recall $>0.14$ and at or above the reference ROC curves throughout on the
shared 2,178-row sample (USA-agency 30-kt labels, 2020--2024).} The shared
sample intersects the 3,004 frozen holdout records with the RITCF
CMA-wind/frame/storm eligibility records, then retains rows with all 12 SHIPS
predictors finite. It contains 209 RI events (base rate 9.60\%), whereas the
full holdout has a base rate of 7.39\%. All models are evaluated on the same
rows under the same USA-agency 30-kt labels. RITCF curves and public 6/10
points use best-AUC checkpoints; inner-F1 operating markers use the stated
validation sweep.}
\label{fig:s_curves}
\end{figure}

\begin{figure}[!htbp]
\centering
\includegraphics[width=0.9\textwidth]{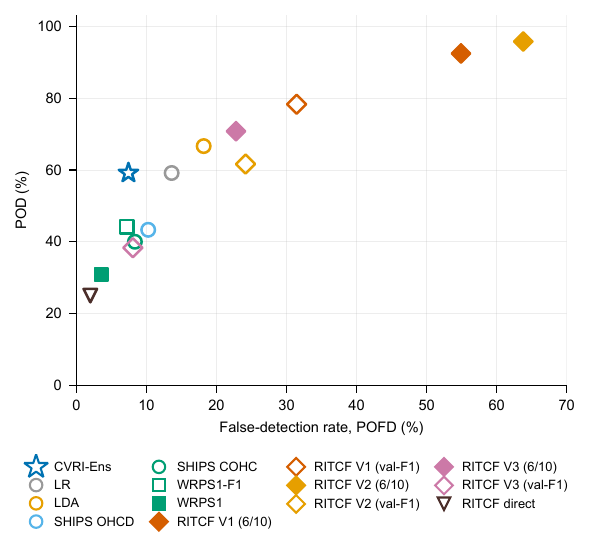}
\caption{\textbf{Under CMA 30-kt labels on the same 2,178 rows, CVRI-Ens
retains the highest CSI and the highest POD at 9.4\% FARate.} All models
are re-scored from frozen predictions under the CMA labels. Open markers
use thresholds re-derived on inner-period data under those labels; filled
markers use public or locked rules. The 9.4\% FARate POD values are
interpolated from the empirical POD--FARate curves; RITCF markers use
best-AUC checkpoints.}
\label{fig:s_cma}
\end{figure}

\begin{figure}[!htbp]
\centering
\includegraphics[width=0.75\textwidth]{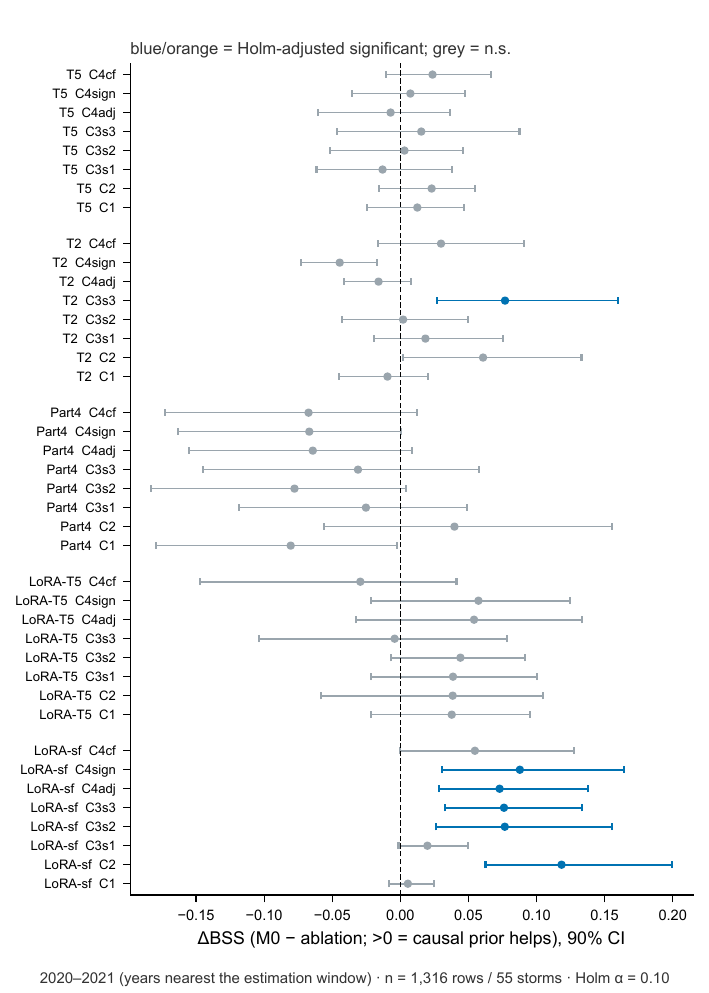}
\caption{\textbf{On the test years nearest the estimation window
(2020--2021), all five members benefit from the graph pathway and six of
the 40 contrasts are Holm-significant, all favouring the prior.} Ablation
deltas ($\Delta$BSS $=$ full member (M0) $-$ ablated member; positive
values favour the prior) on the 1,316-row 2020--2021 subset (55 storms),
with storm-cluster bootstrap 90\% confidence intervals and Holm
adjustment within each member's eight-condition family ($\alpha=0.10$);
blue and orange mark contrasts that survive Holm adjustment, grey do not. C1 removes all
causal losses, C2 disables the graph neural network (GNN) branch, C3s1--3
replace the prior graph with shuffles that preserve each node's degree in
the environmental subgraph while holding loss weights fixed, and C4 retains a
single loss term. The full-period contrasts are listed in Supplementary Table S5.}
\label{fig:s_forest}
\end{figure}

\begin{figure}[!htbp]
\centering
\includegraphics[width=0.9\textwidth]{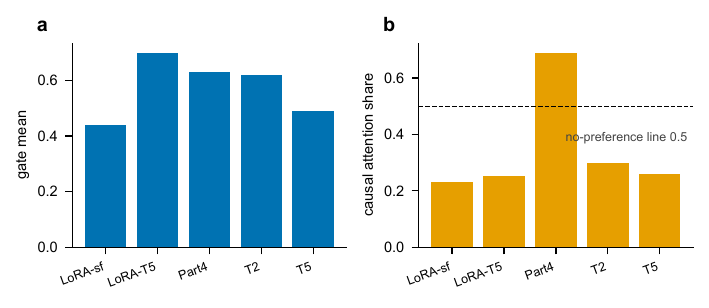}
\caption{\textbf{Members differ in fusion-gate activation and in the share
of cross-attention assigned to the prior.} \textbf{a}, Mean fusion-gate
activation per member on the 3,004 test rows. \textbf{b}, Head-averaged
share of cross-attention assigned to the causal token (reference 0.5 for two
context tokens).}
\label{fig:s_gate}
\end{figure}

\clearpage
\bibliographystyle{plainnat}
\bibliography{references}

%% file: sm_tables.tex
% ---------------------------------------------------------------------------
% Supplementary Tables S1-S9 (this file is \input by supplementary.tex).
% Generated by wpac_causal_phase2/scripts/make_sm_tables.py -- do not edit by hand.
% Requires \usepackage{booktabs} and \usepackage{longtable} in the preamble;
% S1, S2 and S5 are longtables and may break across pages.
% Every number is read from a frozen artefact; the '% source:' line under each
% table names its input file.
% ---------------------------------------------------------------------------

% ===== Supplementary Table S1: Average treatment effects (full table) =====
{\scriptsize
\setlength{\tabcolsep}{2.5pt}
\begin{longtable}{@{}lrrrrrrrrp{3.1cm}@{}}
\caption{Ocean heat content provides the largest positive graph-adjusted effects on 24-h intensity change, and only three of the twenty contrasts have a bootstrap confidence interval that crosses zero. Graph-adjusted effect estimates for the ten environmental treatments on 24-h intensity change (DV24, kt) and rapid intensification (RI, conditional log-odds) in the fit-period (2000--2017) panel (n~=~12,647), with a 95\% percentile bootstrap interval (200 storm-level resamples), the E-value at the estimate and at the confidence limit nearest the null, the bootstrap p-value, the number of the four refutation tests passed, and the backdoor adjustment set implied by the locked DAG (``none'' = empty set).}
\label{tab:S1}\\
\toprule
\textbf{Treatment} & \textbf{Outcome} & \textbf{Effect} & \textbf{CI low} & \textbf{CI high} & \textbf{E-value} & \textbf{E-value CI limit} & $\boldsymbol{p}_{\mathrm{boot}}$ & \textbf{Refut.} & \textbf{Adjustment set} \\
\midrule
\endfirsthead
\multicolumn{10}{@{}l}{\footnotesize\itshape (continued)}\\
\toprule
\textbf{Treatment} & \textbf{Outcome} & \textbf{Effect} & \textbf{CI low} & \textbf{CI high} & \textbf{E-value} & \textbf{E-value CI limit} & $\boldsymbol{p}_{\mathrm{boot}}$ & \textbf{Refut.} & \textbf{Adjustment set} \\
\midrule
\endhead
\bottomrule
\endfoot
SHRD & DV24 & $-$4.788 & $-$5.387 & $-$4.147 & 1.789 & 1.705 & 0.005 & 3/4 & VORT \\
SHRD & RI & $-$0.729 & $-$0.872 & $-$0.595 & 3.386 & 2.891 & 0.005 & 2/4 & VORT \\
RHLO & DV24 & 2.791 & 2.045 & 3.485 & 1.525 & 1.423 & 0.005 & 2/4 & W500 \\
RHLO & RI & 0.112 & 0.003 & 0.211 & 1.453 & 1.059 & 0.040 & 1/4 & W500 \\
VORT & DV24 & $-$3.825 & $-$4.356 & $-$3.237 & 1.662 & 1.585 & 0.005 & 3/4 & LAT \\
VORT & RI & $-$0.152 & $-$0.277 & $-$0.041 & 1.567 & 1.240 & 0.010 & 2/4 & LAT \\
PV & DV24 & $-$5.062 & $-$5.619 & $-$4.499 & 1.826 & 1.751 & 0.005 & 3/4 & SHRD, W500, SST\_IC \\
PV & RI & 0.089 & $-$0.121 & 0.292 & 1.389 & 1.000 & 0.430 & 2/4 & SHRD, W500, WARM\_CORE \\
D200 & DV24 & 4.945 & 4.182 & 5.502 & 1.810 & 1.709 & 0.005 & 3/3 & none \\
D200 & RI & 0.300 & 0.166 & 0.424 & 1.953 & 1.601 & 0.005 & 2/4 & W500, RHLO \\
WARM\_CORE & DV24 & $-$4.763 & $-$5.254 & $-$4.233 & 1.786 & 1.716 & 0.005 & 3/4 & W500, SST\_IC \\
WARM\_CORE & RI & 0.129 & 0.053 & 0.217 & 1.500 & 1.279 & 0.005 & 2/4 & W500, SST\_IC \\
MFC & DV24 & 0.314 & $-$0.306 & 0.841 & 1.135 & 1.000 & 0.310 & 3/4 & W500, VORT \\
MFC & RI & 0.070 & $-$0.038 & 0.174 & 1.332 & 1.000 & 0.210 & 2/4 & W500, VORT, RHLO \\
W500 & DV24 & $-$3.638 & $-$4.116 & $-$3.108 & 1.638 & 1.567 & 0.005 & 3/4 & SHRD \\
W500 & RI & $-$0.251 & $-$0.352 & $-$0.138 & 1.837 & 1.530 & 0.005 & 2/4 & SHRD \\
OHC\_IC & DV24 & 5.886 & 4.473 & 7.194 & 1.935 & 1.748 & 0.005 & 2/4 & OHC\_OA \\
OHC\_IC & RI & 0.745 & 0.427 & 1.003 & 3.279 & 2.301 & 0.005 & 2/4 & OHC\_OA \\
OHC\_OA & DV24 & 6.977 & 6.061 & 7.819 & 2.084 & 1.959 & 0.005 & 2/4 & SHRD, W500, SST\_IC, SST\_OA \\
OHC\_OA & RI & 0.668 & 0.525 & 0.836 & 3.026 & 2.585 & 0.005 & 2/4 & W500, WARM\_CORE, VORT, SST\_OA, MFC \\
\end{longtable}}
\par\smallskip\noindent{\footnotesize\raggedright DV24 ($\Delta V_{24}$), 24-h intensity change; RI, rapid intensification; ATE, average treatment effect. Effects are estimated on the fit period (2000--2017) only; bootstrap intervals are conditional on the locked identification DAG. E-value is the minimum strength of association that an unmeasured confounder would need to have with both treatment and outcome to explain the effect away; the E-value CI limit is the same quantity evaluated at the confidence limit closest to zero, and equals 1.000 when that limit already reaches the null. Refut., number of refutation tests passed (four, or three where the adjustment set is empty); these tests use LPM-scale reference effects for both outcomes. Treatments: SHRD, 850--200 hPa vertical wind shear; RHLO, low-level relative humidity; VORT, 850 hPa relative vorticity; PV, potential vorticity; D200, 200 hPa divergence; WARM\_CORE, warm-core 200-hPa temperature anomaly (inner minus outer); MFC, moisture flux convergence; W500, 500 hPa pressure velocity (negative values denote ascent); OHC\_IC, inner-core ocean heat content; OHC\_OA, outer-area ocean heat content. D200 is the only treatment without a directed path to DV24 in the identification DAG, and its RI row relies on an inserted structural edge; both values are reported as associations rather than identified ATEs. D200 is excluded from the trusted signs by a protocol-locked rule. RI estimates are conditional log-odds per SD (binomial GLM), not a marginal ATE.\par}
% source: wpac_causal_phase1/data/processed_masked/holdout_fit0017/prior/ate_results.csv

% ===== Supplementary Table S2: 22-arm inner ranking =====
\clearpage
{\scriptsize
\begin{longtable}{@{}rllrrrrr@{}}
\caption{Ranking the 22 arms by their worse inner year selects five arms from two of the four families, and the resulting uniform-weight ensemble scores a pooled inner BSS of 0.332, above the highest single-arm pooled inner BSS (0.316). Inner-year selection table: per-year inner BSS for 2018 and 2019, the worst of the two years used as the ranking key, the inner-year F1 threshold of each selected member and whether the arm entered the production ensemble.}
\label{tab:S2}\\
\toprule
\textbf{Rank} & \textbf{Arm} & \textbf{Family} & \textbf{Threshold} & \textbf{BSS 2018} & \textbf{BSS 2019} & \textbf{Worst year} & \textbf{Selected} \\
\midrule
\endfirsthead
\multicolumn{8}{@{}l}{\footnotesize\itshape (continued)}\\
\toprule
\textbf{Rank} & \textbf{Arm} & \textbf{Family} & \textbf{Threshold} & \textbf{BSS 2018} & \textbf{BSS 2019} & \textbf{Worst year} & \textbf{Selected} \\
\midrule
\endhead
\bottomrule
\endfoot
1 & \texttt{ens\_lbB\_partial4\_sf} & lingbot\_sf & 0.150 & 0.260 & 0.314 & \textbf{0.2604} & \textbf{yes} \\
2 & \texttt{ens\_lbB\_lora16\_temp5} & lingbot\_mf & 0.250 & 0.383 & 0.247 & \textbf{0.2466} & \textbf{yes} \\
3 & \texttt{ens\_lbB\_temp2} & lingbot\_mf & 0.225 & 0.322 & 0.231 & \textbf{0.2306} & \textbf{yes} \\
4 & \texttt{ens\_lbB\_lora16\_sf} & lingbot\_sf & 0.200 & 0.268 & 0.227 & \textbf{0.2275} & \textbf{yes} \\
5 & \texttt{ens\_lbB\_temp5\_slim} & lingbot\_mf & 0.250 & 0.299 & 0.227 & \textbf{0.2266} & \textbf{yes} \\
6 & \texttt{ens\_lbB\_lpft4\_sf} & lingbot\_sf & -- & 0.256 & 0.219 & 0.2195 & no \\
7 & \texttt{ens\_lingbotB} & lingbot\_sf & -- & 0.301 & 0.212 & 0.2123 & no \\
8 & \texttt{ens\_lbL\_lpft4\_sf} & lingbot\_sf & -- & 0.338 & 0.209 & 0.2086 & no \\
9 & \texttt{ens\_lbB\_temp5\_attnpool} & lingbot\_mf & -- & 0.319 & 0.208 & 0.2082 & no \\
10 & \texttt{ens\_lbB\_temp3} & lingbot\_mf & -- & 0.320 & 0.207 & 0.2073 & no \\
11 & \texttt{ens\_lbB\_temp5\_diff} & lingbot\_mf & -- & 0.321 & 0.207 & 0.2070 & no \\
12 & \texttt{ens\_ep40ft} & mae\_sf & -- & 0.264 & 0.206 & 0.2064 & no \\
13 & \texttt{ens\_lb\_temp5} & lingbot\_mf & -- & 0.291 & 0.202 & 0.2023 & no \\
14 & \texttt{ens\_lbL\_lora16\_sf} & lingbot\_sf & -- & 0.322 & 0.201 & 0.2014 & no \\
15 & \texttt{ens\_ep40\_srv} & mae\_sf & -- & 0.300 & 0.199 & 0.1992 & no \\
16 & \texttt{ens\_ep40} & mae\_sf & -- & 0.298 & 0.198 & 0.1976 & no \\
17 & \texttt{ens\_lingbotL\_temp5} & lingbot\_mf & -- & 0.366 & 0.193 & 0.1932 & no \\
18 & \texttt{ens\_v1} & mae\_sf & -- & 0.283 & 0.180 & 0.1799 & no \\
19 & \texttt{ens\_lbB\_temp5\_diffslim} & lingbot\_mf & -- & 0.285 & 0.176 & 0.1763 & no \\
20 & \texttt{ens\_lingbot} & lingbot\_sf & -- & 0.335 & 0.174 & 0.1744 & no \\
21 & \texttt{ens\_lingbotL} & lingbot\_sf & -- & 0.287 & 0.171 & 0.1709 & no \\
22 & \texttt{ens\_ep40\_temp5} & mae\_mf & -- & 0.250 & 0.154 & 0.1540 & no \\
\end{longtable}}
\par\smallskip\noindent{\footnotesize\raggedright Selection rule (protocol\_locks/selection.json): per-year inner BSS for 2018 and 2019, ranked by the worse year, tie-broken by inner mean, then family order and arm name. Family gate (8): a gated-family champion receives a protected slot only if it ranks within the overall top 8 by two-year-worst inner BSS; otherwise the family is not admitted, and remaining slots are filled by overall rank. Both MAE champions fell outside the gate (ranks 12 and 22), so the five highest-ranked arms form the ensemble. K~=~5, uniform weights, ensemble threshold from the inner F1 optimum. Thresholds are shown only for the five selected members. Arms are LingBot/MAE checkpoints differing in fine-tuning mode, LoRA rank, temporal pooling and input channels. For the historical run IDs ens\_ep40ft and ens\_ep40\_temp5, ep40 is a run label, not an epoch-40 checkpoint identifier; the visual branch loads mae\_vit\_s16\_best.pt. Short names used in the main text, figures and Tables S5--S6: LoRA-sf = ens\_lbB\_lora16\_sf, LoRA-T5 = ens\_lbB\_lora16\_temp5, Part4 = ens\_lbB\_partial4\_sf, T2 = ens\_lbB\_temp2, T5 = ens\_lbB\_temp5\_slim.\par}
% source: wpac_causal_phase2/protocol_locks/selection.csv; wpac_causal_phase2/protocol_locks/selection.json

% ===== Supplementary Table S3: Per-year metrics, 5 models x 2020-2024 =====
\begin{table}[htbp]
\centering
\scriptsize
\caption{CVRI-Ens has the highest BSS$_{\mathrm{dev}}$ in every one of the five test years and the highest CSI in four of them (SHIPS-12 OHCD leads on CSI in 2020 alone), and both SHIPS-12 variants stay positively skilled throughout while the uncalibrated LR baseline does not. Verification metrics per test year (n = 3,004 rows in total; pooled row at the foot of each block) under the shared evaluation protocol: identical rows, identical RI label, fixed development climatology (2000--2019 RI rate 0.0791), and the operational threshold taken as the inner-year F1 optimum for each model.}
\label{tab:S3}
\begin{tabular}{@{}lrrrrrr@{}}
\toprule
\textbf{Year} & \textbf{\textit{n}} & \textbf{BSS$_{\mathrm{dev}}$} & \textbf{POD} & \textbf{POFD} & \textbf{FARatio} & \textbf{CSI} \\
\midrule
\multicolumn{7}{@{}l}{\textit{CVRI-Ens (Top-5)}}\\
2020 & 559 & 0.276 & 0.674 & 0.097 & 0.633 & 0.312 \\
2021 & 757 & 0.298 & 0.659 & 0.073 & 0.642 & 0.302 \\
2022 & 571 & 0.159 & 0.639 & 0.099 & 0.697 & 0.258 \\
2023 & 606 & 0.287 & 0.660 & 0.076 & 0.545 & 0.368 \\
2024 & 511 & 0.252 & 0.761 & 0.142 & 0.653 & 0.312 \\
\textbf{2020--24} & \textbf{3,004} & \textbf{0.259} & \textbf{0.680} & \textbf{0.095} & \textbf{0.635} & \textbf{0.311} \\
\multicolumn{7}{@{}l}{\textit{SHIPS-12 LDA (COHC)}}\\
2020 & 559 & 0.200 & 0.419 & 0.041 & 0.538 & 0.281 \\
2021 & 757 & 0.134 & 0.477 & 0.063 & 0.682 & 0.236 \\
2022 & 571 & 0.036 & 0.167 & 0.039 & 0.778 & 0.105 \\
2023 & 606 & 0.076 & 0.396 & 0.103 & 0.731 & 0.191 \\
2024 & 511 & 0.074 & 0.239 & 0.052 & 0.686 & 0.157 \\
\textbf{2020--24} & \textbf{3,004} & \textbf{0.105} & \textbf{0.347} & \textbf{0.060} & \textbf{0.686} & \textbf{0.197} \\
\multicolumn{7}{@{}l}{\textit{SHIPS-12 LDA (OHCD)}}\\
2020 & 559 & 0.201 & 0.581 & 0.064 & 0.569 & 0.329 \\
2021 & 757 & 0.112 & 0.659 & 0.115 & 0.739 & 0.230 \\
2022 & 571 & 0.006 & 0.250 & 0.084 & 0.833 & 0.111 \\
2023 & 606 & 0.063 & 0.453 & 0.125 & 0.742 & 0.197 \\
2024 & 511 & 0.086 & 0.391 & 0.073 & 0.654 & 0.225 \\
\textbf{2020--24} & \textbf{3,004} & \textbf{0.095} & \textbf{0.473} & \textbf{0.095} & \textbf{0.715} & \textbf{0.216} \\
\multicolumn{7}{@{}l}{\textit{LDA (21 panel predictors)}}\\
2020 & 559 & 0.220 & 0.744 & 0.238 & 0.794 & 0.193 \\
2021 & 757 & 0.096 & 0.818 & 0.251 & 0.833 & 0.161 \\
2022 & 571 & $-$0.014 & 0.528 & 0.280 & 0.888 & 0.102 \\
2023 & 606 & 0.075 & 0.755 & 0.246 & 0.773 & 0.212 \\
2024 & 511 & 0.099 & 0.761 & 0.232 & 0.755 & 0.227 \\
\textbf{2020--24} & \textbf{3,004} & \textbf{0.098} & \textbf{0.730} & \textbf{0.250} & \textbf{0.811} & \textbf{0.176} \\
\multicolumn{7}{@{}l}{\textit{LR (uncalibrated)}}\\
2020 & 559 & $-$2.028 & 0.721 & 0.167 & 0.735 & 0.240 \\
2021 & 757 & $-$2.708 & 0.682 & 0.175 & 0.806 & 0.178 \\
2022 & 571 & $-$2.576 & 0.389 & 0.166 & 0.864 & 0.112 \\
2023 & 606 & $-$1.694 & 0.660 & 0.177 & 0.737 & 0.232 \\
2024 & 511 & $-$1.519 & 0.739 & 0.176 & 0.707 & 0.266 \\
\textbf{2020--24} & \textbf{3,004} & \textbf{$-$2.074} & \textbf{0.649} & \textbf{0.173} & \textbf{0.769} & \textbf{0.205} \\
\bottomrule
\end{tabular}
\par\smallskip\noindent{\footnotesize\raggedright BSS$_{\mathrm{dev}}$, Brier skill score against the fixed development climatology (2000--2019 RI rate 0.0791), identical for every model and year; POD, probability of detection; POFD, probability of false detection (equals FARate); FARatio, false-alarm ratio; CSI, critical success index. Individual-year counts (511--757 rows) limit the precision of year-to-year contrasts; the pooled row is the primary basis for comparison. Outside the uncalibrated LR baseline, whose BSS$_{\mathrm{dev}}$ is negative in all five years, the only negative year-cell is the 21-predictor panel LDA in 2022 ($-$0.014).\par}
\end{table}
% source: wpac_causal_phase2/data/analysis/eval/{main_2024,baseline_lr,baseline_lda,baseline_ships12_lda,baseline_ships12_lda_cohc}.json

% ===== Supplementary Table S4: Ensemble-size sensitivity (K4/Top-5/K6/all-22) =====
\begin{table}[htbp]
\centering
\scriptsize
\caption{K6 attains BSS$_{\mathrm{dev}}$ 0.279, compared with 0.259 for K4 and Top-5, while all-22 attains 0.244. Full metric panel for the four ensemble sizes evaluated on the same 3,004 test rows under the locked protocol; members enter with uniform weights and each ensemble carries its own inner-year F1 threshold.}
\label{tab:S4}
\begin{tabular}{@{}lrrrrrrrrr@{}}
\toprule
\textbf{Ensemble} & \textbf{\textit{K}} & \textbf{\textit{n}} & \textbf{BSS$_{\mathrm{dev}}$} & \textbf{ROC AUC} & \textbf{PR AUC} & \textbf{POD} & \textbf{POFD} & \textbf{FARatio} & \textbf{CSI} \\
\midrule
K4 (Top-4) & 4 & 3,004 & 0.259 & 0.898 & 0.513 & 0.667 & 0.087 & 0.621 & 0.319 \\
Top-5 (main) & 5 & 3,004 & 0.259 & 0.900 & 0.518 & 0.680 & 0.095 & 0.635 & 0.311 \\
K6 (Top-6) & 6 & 3,004 & 0.279 & 0.902 & 0.527 & 0.662 & 0.076 & 0.589 & 0.339 \\
all-22 & 22 & 3,004 & 0.244 & 0.905 & 0.509 & 0.743 & 0.128 & 0.684 & 0.285 \\
\bottomrule
\end{tabular}
\par\smallskip\noindent{\footnotesize\raggedright K, number of members; the K6 arm adds the sixth-ranked inner arm (ens\_lbB\_lpft4\_sf) to the protocol-locked Top-5, and all-22 pools every trained arm. K4/K6 and all-22 are sensitivity analyses: the protocol-locked Top-5 remains the primary ensemble even where K6 scores higher.\par}
\end{table}
% source: wpac_causal_phase2/data/analysis/eval/{k4_2024,main_2024,k6_2024,all22_2024}.json
\clearpage

% ===== Supplementary Table S5: 40-entity ablation panel =====
{\scriptsize
\begin{longtable}{@{}llrrrc@{}}
\caption{Ten of the 40 member-by-condition contrasts survive Holm correction and all ten fall in three of the five members, while the protocol-locked C2 contrast (graph branch removed) reaches Holm significance in LoRA-sf. Full ablation panel: delta BSS (M0 minus ablation, positive = the causal prior helps) with 90\% storm-cluster bootstrap intervals and Holm-adjusted p-values, for five ensemble members each ablated in eight conditions on the 3,004 test rows.}
\label{tab:S5}\\
\toprule
\textbf{Member} & \textbf{Condition} & $\boldsymbol{\Delta}$\textbf{BSS} & \textbf{90\% CI} & $\boldsymbol{p}_{\mathrm{Holm}}$ & \textbf{Significant} \\
\midrule
\endfirsthead
\multicolumn{6}{@{}l}{\footnotesize\itshape (continued)}\\
\toprule
\textbf{Member} & \textbf{Condition} & $\boldsymbol{\Delta}$\textbf{BSS} & \textbf{90\% CI} & $\boldsymbol{p}_{\mathrm{Holm}}$ & \textbf{Significant} \\
\midrule
\endhead
\bottomrule
\endfoot
\texttt{Part4} & C1 & $-$0.093 & $[$$-$0.160, $-$0.038$]$ & 0.024 & \textbf{yes} \\
\texttt{Part4} & C2 & $-$0.014 & $[$$-$0.079, 0.053$]$ & 0.681 & no \\
\texttt{Part4} & C3s1 & $-$0.076 & $[$$-$0.148, $-$0.017$]$ & 0.174 & no \\
\texttt{Part4} & C3s2 & $-$0.071 & $[$$-$0.142, $-$0.014$]$ & 0.174 & no \\
\texttt{Part4} & C3s3 & $-$0.059 & $[$$-$0.142, 0.007$]$ & 0.292 & no \\
\texttt{Part4} & C4adj & $-$0.072 & $[$$-$0.135, $-$0.018$]$ & 0.174 & no \\
\texttt{Part4} & C4sign & $-$0.089 & $[$$-$0.155, $-$0.036$]$ & 0.035 & \textbf{yes} \\
\texttt{Part4} & C4cf & $-$0.062 & $[$$-$0.128, $-$0.007$]$ & 0.186 & no \\
\texttt{LoRA-T5} & C1 & 0.028 & $[$$-$0.012, 0.073$]$ & 1.000 & no \\
\texttt{LoRA-T5} & C2 & $-$0.026 & $[$$-$0.082, 0.024$]$ & 1.000 & no \\
\texttt{LoRA-T5} & C3s1 & 0.015 & $[$$-$0.033, 0.066$]$ & 1.000 & no \\
\texttt{LoRA-T5} & C3s2 & 0.038 & $[$0.000, 0.075$]$ & 0.700 & no \\
\texttt{LoRA-T5} & C3s3 & 0.010 & $[$$-$0.039, 0.055$]$ & 1.000 & no \\
\texttt{LoRA-T5} & C4adj & 0.035 & $[$$-$0.015, 0.084$]$ & 1.000 & no \\
\texttt{LoRA-T5} & C4sign & 0.051 & $[$0.004, 0.099$]$ & 0.632 & no \\
\texttt{LoRA-T5} & C4cf & $-$0.008 & $[$$-$0.064, 0.040$]$ & 1.000 & no \\
\texttt{T2} & C1 & $-$0.026 & $[$$-$0.051, $-$0.004$]$ & 0.225 & no \\
\texttt{T2} & C2 & 0.032 & $[$$-$0.002, 0.067$]$ & 0.436 & no \\
\texttt{T2} & C3s1 & 0.001 & $[$$-$0.021, 0.027$]$ & 1.000 & no \\
\texttt{T2} & C3s2 & $-$0.004 & $[$$-$0.036, 0.025$]$ & 1.000 & no \\
\texttt{T2} & C3s3 & 0.068 & $[$0.031, 0.114$]$ & 0.014 & \textbf{yes} \\
\texttt{T2} & C4adj & $-$0.024 & $[$$-$0.045, $-$0.007$]$ & 0.138 & no \\
\texttt{T2} & C4sign & $-$0.039 & $[$$-$0.057, $-$0.023$]$ & 0.004 & \textbf{yes} \\
\texttt{T2} & C4cf & 0.018 & $[$$-$0.017, 0.062$]$ & 1.000 & no \\
\texttt{LoRA-sf} & C1 & 0.007 & $[$$-$0.003, 0.018$]$ & 0.518 & no \\
\texttt{LoRA-sf} & C2 & 0.080 & $[$0.049, 0.119$]$ & 0.004 & \textbf{yes} \\
\texttt{LoRA-sf} & C3s1 & 0.010 & $[$$-$0.004, 0.024$]$ & 0.518 & no \\
\texttt{LoRA-sf} & C3s2 & 0.062 & $[$0.030, 0.101$]$ & 0.004 & \textbf{yes} \\
\texttt{LoRA-sf} & C3s3 & 0.055 & $[$0.028, 0.088$]$ & 0.010 & \textbf{yes} \\
\texttt{LoRA-sf} & C4adj & 0.054 & $[$0.024, 0.091$]$ & 0.012 & \textbf{yes} \\
\texttt{LoRA-sf} & C4sign & 0.066 & $[$0.035, 0.103$]$ & 0.004 & \textbf{yes} \\
\texttt{LoRA-sf} & C4cf & 0.050 & $[$0.017, 0.093$]$ & 0.030 & \textbf{yes} \\
\texttt{T5} & C1 & $-$0.001 & $[$$-$0.020, 0.018$]$ & 1.000 & no \\
\texttt{T5} & C2 & 0.005 & $[$$-$0.018, 0.024$]$ & 1.000 & no \\
\texttt{T5} & C3s1 & $-$0.022 & $[$$-$0.048, 0.002$]$ & 1.000 & no \\
\texttt{T5} & C3s2 & $-$0.020 & $[$$-$0.052, 0.005$]$ & 1.000 & no \\
\texttt{T5} & C3s3 & 0.002 & $[$$-$0.033, 0.036$]$ & 1.000 & no \\
\texttt{T5} & C4adj & $-$0.016 & $[$$-$0.040, 0.005$]$ & 1.000 & no \\
\texttt{T5} & C4sign & 0.019 & $[$$-$0.002, 0.042$]$ & 1.000 & no \\
\texttt{T5} & C4cf & 0.008 & $[$$-$0.010, 0.029$]$ & 1.000 & no \\
\end{longtable}}
\par\smallskip\noindent{\footnotesize\raggedright Cluster bootstrap over 131 storms, 2,000 resamples, 90\% CI; Holm correction within each member's eight-condition family at $\alpha = 0.10$, with adjustment confined to each member's block. Conditions: C1, all causal losses removed; C2, causal losses and the graph branch removed; C3s1--C3s3, prior shuffled with each node's degree preserved in the environmental subgraph and loss weights unchanged, with three fixed seeds; C4adj, adjacency term only; C4sign, sign term only; C4cf, counterfactual term only. ``yes'' marks Holm rejection; direction of the estimate is given by the sign of delta BSS.\par}
% source: wpac_causal_phase2/data/analysis/eval/M0_ens_lbB_*.json (paired_bootstrap)

% ===== Supplementary Table S6: Gate decomposition (5 members) =====
\begin{table}[htbp]
\centering
\scriptsize
\caption{Members route different shares of their cross-attention to the causal graph token (causal share 0.230 to 0.689; gate mean 0.438 to 0.698); only one member's causal share sits above the 0.5 no-preference line. Gate decomposition of the five ensemble members over the 3,004 test rows: mean fusion gate, head-averaged causal-token attention share, its deviation from the two-token no-preference line, the resulting causal influence weight, and the RI-minus-non-RI gate contrast.}
\label{tab:S6}
\begin{tabular}{@{}lrrrrrrr@{}}
\toprule
\textbf{Member} & \textbf{gate mean} & \textbf{causal share} & \textbf{share$-$0.5} & \textbf{gate$\times$share} & \textbf{gate RI} & \textbf{gate non-RI} & $\boldsymbol{\Delta}$\textbf{gate} \\
\midrule
\texttt{Part4} & 0.629 & 0.689 & 0.189 & 0.426 & 0.665 & 0.626 & 0.039 \\
\texttt{T2} & 0.620 & 0.299 & $-$0.201 & 0.188 & 0.677 & 0.615 & 0.062 \\
\texttt{T5} & 0.488 & 0.260 & $-$0.240 & 0.141 & 0.506 & 0.487 & 0.020 \\
\texttt{LoRA-T5} & 0.698 & 0.251 & $-$0.249 & 0.175 & 0.666 & 0.701 & $-$0.035 \\
\texttt{LoRA-sf} & 0.438 & 0.230 & $-$0.270 & 0.091 & 0.414 & 0.440 & $-$0.026 \\
\bottomrule
\end{tabular}
\par\smallskip\noindent{\footnotesize\raggedright Per-row means over the 3,004 test rows. gate\_mean is averaged over the 256 dimensions of the sigmoid gate on the fused environmental and causal attention output and over the test rows; causal\_share is the head-averaged cross-attention share of the causal GNN token, where 0.5 is the no-preference line with two tokens; gate$\times$share is the mean of the per-row product, used as a descriptive causal influence weight. All columns are descriptive mechanism summaries and carry no test of a mediation claim.\par}
\end{table}
% source: wpac_causal_phase2/data/analysis/gate_participation.csv

% ===== Supplementary Table S7: Sample-size census (8 definitions) =====
\begin{table}[htbp]
\centering
\scriptsize
\caption{Eight sample frames are used in this study and are not interchangeable. The five in-house frames (3,004, 2,892, 2,178, 1,316 and 949 rows) are subsets of the official test table; 2,178 and 1,316 overlap, and 949 is their common subset. The two reconstructed RITCF frames (2,385 and 1,057 rows) are used for the protocol reconstructions, and the consistent-comparison panels (2,178 and 949 rows) are their intersections with our frozen rows. The 1,149-row frame is the external anchor reported by \citet{wang2025ritcf}. Sample-size census, with the use to which each row count belongs.}
\label{tab:S7}
\begin{tabular}{@{}lrr@{}}
\toprule
\textbf{Sample definition} & \textbf{\textit{n}} & \textbf{Use} \\
\midrule
Official test set (main table) & 3,004 & All headline numbers \\
WRPS1 same-sample subset & 2,892 & WRPS1 delta-BSS bootstrap \\
RITCF reconstruction, CMA 30-kt frame & 2,385 & RITCF CMA 30-kt protocol \\
\textbf{Panel A intersection} & \textbf{2,178} & \textbf{Consistent-comparison panel, our label standard} \\
2020--2021 subset & 1,316 & Two-year supplementary contrast \\
RITCF as published & 1,149 & Their reported operating point (external anchor) \\
RITCF reconstruction, CMA 25.27-kt frame & 1,057 & Independent reconstruction frame \\
\textbf{Panel B intersection} & \textbf{949} & \textbf{Consistent-comparison panel, their label standard} \\
\bottomrule
\end{tabular}
\par\smallskip\noindent{\footnotesize\raggedright Counts are rows surviving frame availability and the all-finite SHIPS-12 filter, where that filter applies; the two reconstructed RITCF frames (2,385 and 1,057 rows) are intersected with our frozen test rows before any comparison (Supplementary Table S8), and 1,149 is the row count of the published operating point, which is used as an external anchor only.\par}
\end{table}
% source: wpac_causal_phase2/docs/RESULTS_HOLDOUT_MASKED_V1.md Sec. 10 (sample-size census)

% ===== Supplementary Table S8: Wind-source label control =====
\begin{table}[htbp]
\centering
\scriptsize
\caption{The two wind sources assign the same RI label to 95.4\% of the 2,178 panel-A rows; 114 of the 215 rows positive under either source coincide (Jaccard 0.530). Panel A compares USA-30 with CMA-30 (209 and 120 RI rows; base rates 0.096 and 0.055), and panel B compares USA-30 with CMA-25.27 (83 and 66; 0.087 and 0.070). RI counts and base rates are shown for both panels; RI-set Jaccard and row-level label agreement are shown for panel A only.}
\label{tab:S8}
\begin{tabular}{@{}lrrrrrrr@{}}
\toprule
\textbf{Panel} & \textbf{\textit{n}} & \textbf{RI USA} & \textbf{RI CMA} & \textbf{Rate USA} & \textbf{Rate CMA} & \textbf{Jaccard} & \textbf{Agreement} \\
\midrule
A (our standard) & 2,178 & 209 & 120 & 0.096 & 0.055 & 0.530 & 0.954 \\
B (their standard) & 949 & 83 & 66 & 0.087 & 0.070 & -- & -- \\
\bottomrule
\end{tabular}
\par\smallskip\noindent{\footnotesize\raggedright Panel A, rows of our frozen test set intersecting the RITCF CMA 30-kt reconstruction and all-finite SHIPS-12 rows (2020--2024); panel B, the CMA 25.27-kt reconstruction frame intersected with our frozen predictions and all-finite SHIPS-12 rows (2020--2021). Base rates, Jaccard and agreement are fractions of the panel rows; Jaccard is computed on the RI-positive rows of panel A and agreement is the fraction of its rows whose binary RI label is identical under the two wind sources. For panel B only the per-source RI counts are reported because the two label sets are not both locked on that frame. The lower CMA base rate is part of the label shift in the cross-label controls.\par}
\end{table}
% source: wpac_causal_phase2/data/analysis/compare/coverage.csv; wpac_causal_phase2/protocol_locks/comparison_secondary.json (wind_source_decomposition); wpac_causal_phase2/docs/RESULTS_HOLDOUT_MASKED_V1.md Sec. 10

% ===== Supplementary Table S9: SHIPS-12 predictor table =====
\begin{table}[htbp]
\centering
\scriptsize
\setlength{\tabcolsep}{4pt}
\caption{Six of the twelve SHIPS predictors are reproduced exactly, one near-exactly and one by interpolation; the remaining substitutions are recorded below. The twelve predictors of Horinouchi et al.\ (2025) as re-derived on our own panel, with the radius of the area average, the source used here and the fidelity label recorded in the baseline documentation.}
\label{tab:S9}
\begin{tabular}{@{}rlp{4.5cm}p{5.0cm}l@{}}
\toprule
\textbf{\#} & \textbf{Predictor} & \textbf{Definition (radius)} & \textbf{Source in this study} & \textbf{Fidelity} \\
\midrule
1 & \texttt{VMAX} & Maximum sustained wind (kt) & Panel VMAX (IBTrACS USA\_WIND, 1-min) & EXACT \\
2 & \texttt{INCV} & 6-h intensity change, t$-$6~h to t0 (kt) & Panel PER (6-h IBTrACS track difference) & EXACT \\
3 & \texttt{POT} & Potential intensity minus VMAX (kt), 0--200 km & tcpyPI 1.4.1 \citep{gilford2021tcpypi} on ERA5 soundings, 0--200 km; Bister--Emanuel algorithm \citep{bister2002potential} & APPROX \\
4 & \texttt{OHCD} & Ocean heat content (kJ~cm$^{-2}$), 0--100 km & GLORYS12V1 daily field, closed window (source\_time $+$ 24~h $\leq$ t0) & BETTER \\
5 & \texttt{SHRD} & 850--200 hPa shear magnitude (kt), 200--800 km & ERA5 u/v annular mean difference & EXACT \\
6 & \texttt{T150} & 150 hPa temperature ($^{\circ}$C), 200--800 km & Log-p extrapolation from 200/300 hPa (no local 150 hPa level) & SUBSTITUTED \\
7 & \texttt{T200} & 200 hPa temperature ($^{\circ}$C), 200--800 km & ERA5 t at 200 hPa & EXACT \\
8 & \texttt{T250} & 250 hPa temperature ($^{\circ}$C), 200--800 km & Log-p interpolation of 200/300 hPa (244.9 hPa) & INTERP \\
9 & \texttt{R000} & 1000 hPa relative humidity (\%), 200--800 km & ERA5 r at 1000 hPa & EXACT \\
10 & \texttt{SDIR} & Standard deviation of IR brightness temperature ($^{\circ}$C), 0--200 km & Paired HURSAT-B1 TBB frame (QC v1 corpus) & NEAR-EXACT \\
11 & \texttt{Z850} & 850 hPa relative vorticity ($10^{-7}$~s$^{-1}$), 0--1000 km & ERA5 u/v central difference, signed dlat & EXACT \\
12 & \texttt{DTL} & Great-circle distance to nearest land (km) & Panel LAND\_DIST (GSHHG coastline) & APPROX \\
\bottomrule
\end{tabular}
\par\smallskip\noindent{\footnotesize\raggedright Radii, units and the area-average operator follow WRPS1 Table 1; all fields are taken at the t0 synoptic time (0-h match within a 3-h tolerance). Fidelity labels: EXACT, direct ERA5/HURSAT/panel equivalent; NEAR-EXACT, close correspondence with a documented source difference; INTERP, log-p interpolation; APPROX, local approximation; SUBSTITUTED, layer unavailable locally and replaced; BETTER, time-varying where WRPS1 uses a climatology. The protocol-locked COHC sensitivity replaces OHCD by a fit-only (2000--2017) month-by-0.5$^{\circ}$ GLORYS climatology; the two ocean variants are reported side by side and never enter the same model.\par}
\end{table}
% source: wpac_causal_phase2/docs/SHIPS12_BASELINE.md Sec. 2 (definitions); wpac_causal_phase2/src/baseline_common.py (locked list)